\documentclass[pdflatex,sn-basic]{sn-jnl}% Basic Springer Nature Reference Style/Chemistry Reference Style
\usepackage{graphicx}%
\graphicspath{{./Images/}}
\usepackage{multirow}%
\usepackage{amsmath,amssymb,amsfonts}%
\usepackage{amsthm}%
\usepackage{mathrsfs}%
\usepackage[title]{appendix}%
\usepackage{xcolor}%
\usepackage{textcomp}%
\usepackage{manyfoot}%
\usepackage{booktabs}%
\usepackage{algorithm}%
\usepackage{algorithmicx}%
\usepackage{algpseudocode}%
\usepackage{listings}%
\theoremstyle{thmstyleone}%
\theoremstyle{thmstyletwo}%

\theoremstyle{thmstylethree}%

\usepackage{subcaption}

\begin{document}

\title[Article Title]{Laryngeal Structure Segmentation in High-Speed Videoendoscopy Using Deep Learning}

%%=============================================================%%
%% GivenName	-> \fnm{Joergen W.}
%% Particle	-> \spfx{van der} -> surname prefix
%% FamilyName	-> \sur{Ploeg}
%% Suffix	-> \sfx{IV}
%% \author*[1,2]{\fnm{Joergen W.} \spfx{van der} \sur{Ploeg} 
%%  \sfx{IV}}\email{iauthor@gmail.com}
%%=============================================================%%

\author[1,2]{\fnm{Sardar Nafis} \sur{Bin Ali}}\email{binsarda@msu.edu}

\author[1]{\fnm{Mohsen} \sur{Zayernouri}}\email{zayern@msu.edu}

\author[2]{\fnm{Dimitar D.} \sur{Deliyski}}\email{ddd@msu.edu}

\author*[2]{\fnm{Maryam} \sur{Naghibolhosseini}}\email{naghib@msu.edu}

\affil[1]{\orgdiv{Department of Mechanical Engineering}, \orgname{Michigan State University}, \orgaddress{\street{428 S. Shaw Lane}, \city{East Lansing}, \postcode{48824}, \state{MI}, \country{USA}}}

\affil*[2]{\orgdiv{Department of Communicative Sciences and Disorders}, \orgname{Michigan State University}, \orgaddress{\street{1026 Red Cedar Rd}, \city{East Lansing}, \postcode{48824}, \state{MI}, \country{USA}}}
\abstract{}

\abstract{Laryngeal high-speed videoendoscopy (HSV) offers an effective means of observing the motion of different laryngeal structures along with vibratory behaviors of the vocal folds under various voicing conditions.
Segmentation of laryngeal tissues enables analysis of different tissue structures and their dynamics, helping characterize the involvement of laryngeal muscles in voice production. Given the large number of HSV frames, automating this task is imperative.
While deep learning–based methods have been implemented in previous studies to segment laryngeal structures, they have not been applied to HSV data during connected speech, which poses significant challenges due to excessive tissue movements and image quality limitations associated with fiberoptic image acquisition. 
The application of deep learning to connected speech data is critical for capturing nonstationary laryngeal behaviors and identifying anomalous patterns associated with voice disorders.
The present study aims to address these gaps by training U-Net models to detect the aryepiglottic folds and arytenoid cartilages, vocal folds, epiglottis, and glottal area, using HSV data from both sustained vowel phonation and connected speech obtained from normophonic and disordered voices.
Image pre-processing techniques, including noise removal and histogram equalization, were applied to improve the quality of the training HSV images and enhance network performance.
Finally, to evaluate the accuracy and reliability of the networks, quantitative performance metrics were used alongside qualitative visual inspection of the test images.
The high performance of the developed networks, with overall accuracies exceeding 95\%, establishes their potential as reliable tools for automated laryngeal image analysis, quantitative characterization of laryngeal dynamics, and future detection of anomalous laryngeal behaviors in clinical settings.}


\maketitle

\section{Introduction}\label{intro}
Voice production arises from vocal folds vibration, governed by the complex interaction between aerodynamic forces and the biomechanical properties of the vocal folds \citeauthor{i1} (\citeyear{i1}). The coordinated actions of laryngeal muscles, together with the structural support provided by the cartilages and ligaments, regulate vocal fold position, tension, and configuration to facilitate proper vibration.
The dynamic behavior of laryngeal structures influence voice production and quality. Characterizing anomalies in laryngeal structure dynamics can facilitate the detection of voice disorders and identification of disorder-specific patterns.

Capturing laryngeal images during various speech stimuli enables comprehensive analysis of the dynamic behaviors of different laryngeal components (\citeauthor{i3}, \citeyear{i3}; \citeauthor{i2}, \citeyear{i2}; \citeauthor{i5}, \citeyear{i5}; \citeauthor{i4}, \citeyear{i4}). Stroboscopic laryngeal imaging, or videostroboscopy, has been widely used in voice research and clinical voice assessment (\citeauthor{i7}, \citeyear{i7}; \citeauthor{i8}, \citeyear{i8}; \citeauthor{i9}, \citeyear{i9}; \citeauthor{i10}, \citeyear{i10}; \citeauthor{i12}, \citeyear{i12}; \citeauthor{i11}, \citeyear{i11}; \citeauthor{i6}, \citeyear{i6}; \citeauthor{i15}, \citeyear{i15}; \citeauthor{i13}, \citeyear{i13}; \citeauthor{i14}, \citeyear{i14}). However, the recording frame rate in videostroboscopy is lower than the vibratory frequency of the vocal folds during voice production (\citeauthor{i16}, \citeyear{i16}). Consequently, it captures images of various phases from different vibration cycles and compiles them to represent a single cycle of vocal fold vibration (\citeauthor{i21}, \citeyear{i21}; \citeauthor{i19}, \citeyear{i19}; \citeauthor{i18}, \citeyear{i18}; \citeauthor{i17}, \citeyear{i17}). Therefore, it can accurately and reliably represent a vibration cycle only when the vibration frequency and the displacements at different vibration phases remain constant within each cycle (\citeauthor{i20}, \citeyear{i20}).

In contrast, laryngeal high-speed videoendoscopy (HSV) has a much higher recording frame rate than the vocal fold vibration frequency. Unlike videostroboscopy, it captures multiple phases of vocal fold vibration within each vibratory cycle. Consequently, HSV can capture aperiodic variations in vocal fold vibration that videostroboscopy fails to detect (\citeauthor{i26}, \citeyear{i26}; \citeauthor{i22}, \citeyear{i22}; \citeauthor{i23}, \citeyear{i23}; \citeauthor{i24}, \citeyear{i24}; \citeauthor{i25}, \citeyear{i25}). As a result, HSV is an effective tool for representing true vocal fold vibration cycles, particularly in voice disorders characterized by significant cycle-to-cycle variability and in non-steady phonation events ( \citeauthor{i31}, \citeyear{i31}; \citeauthor{i28}, \citeyear{i28}; \citeauthor{i30}, \citeyear{i30}; \citeauthor{i29}, \citeyear{i29}; \citeauthor{i27}, \citeyear{i27}). However, even a short voiced segment recorded using HSV contains a large number of frames due to its high frame rate. Analysis of laryngeal dynamics requires identifying and tracking relevant anatomical structures across these frames. Manually performing this analysis poses significant challenges and is time-consuming. Therefore, an automated method is needed to identify different laryngeal structures in each frame to quantify their dynamic behavior.

Several studies implemented different techniques for segmenting various laryngeal structures, with most focusing only on the glottal area. Thresholding is a straightforward segmentation approach in which one or multiple laryngeal structures can be identified by selecting appropriate intensity thresholds. Previous studies have applied image intensity–based thresholding (\cite{i32} and \cite{i33}) and histogram-based thresholding (\cite{i34} and \cite{i23}).
However, this method can lead to pixel misclassification since high pixel intensity variations within the same region may cause some pixels to fall outside the defined threshold range, while pixels from surrounding regions may fall within the range and be incorrectly classified. Therefore, the resulting segmentation is often discontinuous and inaccurate, making the method unreliable. The discontinuity issue can be partially improved by adding morphological operations such as opening, closing, gap filling, and erosion. \citeauthor{i35} (\citeyear{i35}) segmented the glottal region by applying binary thresholding, followed by morphological operations. However, the performance of morphological operations is not consistent across all frames, and they cannot accurately recover regions beyond the boundaries initially identified by thresholding.

\citeauthor{i37} (\citeyear{i37}), \citeauthor{i35} (\citeyear{i35}), \citeauthor{i36} (\citeyear{i36}), and \citeauthor{i38} (\citeyear{i38}) used seeded region-growing techniques for glottal area segmentation. These region-growing methods are highly sensitive to the initial selection of correct seed points, noise, and intensity inhomogeneity.
The watershed transformation has also been implemented to segment the glottal area (\citeauthor{i39}, \citeyear{i39}; \citeauthor{i40}, \citeyear{i40}) and while it can be effective for objects with well-defined boundary gradients, its performance deteriorates when boundaries are weak or blurry.
Other segmentation methods involve energy minimization, such as level set methods (\citeauthor{i42}, \citeyear{i42}; \citeauthor{i43}, \citeyear{i43}; \citeauthor{i41}, \citeyear{i41}) and active contour models (\citeauthor{i44}, \citeyear{i44}; \citeauthor{i45}, \citeyear{i45}; \citeauthor{i49}, \citeyear{i49}; \citeauthor{i48}, \citeyear{i48}; \citeauthor{i47}, \citeyear{i47}; \citeauthor{i46}, \citeyear{i46}; \citeauthor{i50}, \citeyear{i50}).
For glottal area segmentation, active contour models (ACM) are more valuable to implement than level set methods because they require lower computational cost, are numerically more stable, and do not require reinitialization. However, ACM is sensitive to image quality and lighting conditions, where proper initialization becomes difficult.

The image-processing algorithms discussed so far rely on predefined rules and parameters to detect different regions and are often ineffective, as these rules and parameters must be repeatedly adjusted across frames and for each subject. Moreover, these methods are difficult to extend to multiclass segmentation and require complex multi-step procedures, often leading to overlaps between different regions. As a result, these methods do not substantially reduce manual effort, visual inspection, and trial-and-error tuning.

Intricate features of different laryngeal zones can be automatically extracted from HSV frames using convolutional neural networks trained on labeled data. Although labeling and training these networks are initially time-consuming and require manual effort, they significantly reduce the time needed for dynamic analysis of various laryngeal tissues by automating the process after training.
Using deep learning networks, many studies focused on segmenting the glottal area (\citeauthor{i53}, \citeyear{i53}; \citeauthor{i54}, \citeyear{i54}; \citeauthor{i55}, \citeyear{i55}; \citeauthor{i56}, \citeyear{i56}; \citeauthor{i51}, \citeyear{i51}; \citeauthor{i52}, \citeyear{i52}), while \citeauthor{i57} (\citeyear{i57}) and \citeauthor{res1} (\citeyear{res1}) segmented both the glottal area and the vocal folds. The datasets used for training in these studies did not include any instances of running speech and were limited to sustained vowel phonation.
Importantly, characteristics of voice disorders are more pronounced in the intra-cycle variations and non-stationary events of connected speech than in sustained vowel phonation (\citeauthor{i58}, \citeyear{i58}; \citeauthor{i59}, \citeyear{i59}; \citeauthor{i60}, \citeyear{i60}; \citeauthor{i61}, \citeyear{i61}; \citeauthor{i62}, \citeyear{i62}; \citeauthor{i63}, \citeyear{i63}; \citeauthor{i64}, \citeyear{i64}; \citeauthor{i50}, \citeyear{i50}).

\citeauthor{i65} (\citeyear{i65}) proposed an unsupervised hybrid approach that uses K-means clustering followed by ACM to identify glottal edges during connected speech.
Although the hybrid method performed well in detecting vocal fold edges during stationary phases, it had difficulty detecting them during non-stationary phases of connected speech (\citeauthor{i68}, \citeyear{i68}).
To overcome these challenges, they later used a deep neural network, with training images automatically labeled using the hybrid method (\citeauthor{i66}, \citeyear{i66}; \citeauthor{i67}, \citeyear{i67}).
However, this deep learning approach showed reduced accuracy in detecting vocal fold edges when the glottal area was very large (\citeauthor{i70}, \citeyear{i70}).
This problem was addressed in several studies (\citeauthor{i68}, \citeyear{i68}; \citeauthor{i69}, \citeyear{i69}; \citeauthor{i70}, \citeyear{i70}; \citeauthor{i71}, \citeyear{i71}) by using manually annotated labels to train deep learning networks. These studies using manual labeling of training data during several instances of connected speech demonstrated promising performance in analyzing dynamics of vocal folds in normophonic and disordered subjects.

The current study aims to extend these studies beyond glottal area segmentation by identifying and segmenting multiple laryngeal structures during both sustained vowel phonation and connected speech in HSV data. To our knowledge, only one study has focused on the segmentation of multiple laryngeal structures \cite{res2}.
\cite{res2} trained a modified YOLOv8n-seg-based machine-learning algorithm to detect several regions, including the vocal folds, glottal area, epiglottis, and five additional regions within the mouth and vocal tract, using frames extracted from videolaryngoscopy recordings of patients with voice disorders.
However, the study did not use HSV, limiting its ability to capture laryngeal structures at the high temporal resolution needed to characterize rapid vocal fold and laryngeal dynamics. Moreover, important laryngeal landmarks, including the aryepiglottic folds and arytenoids, were not segmented.
The current study is designed to address the gaps in existing studies by (a) incorporating identification and segmentation of multiple important laryngeal zones in HSV data, including the vocal folds, glottal area, epiglottis, aryepiglottic folds and arytenoid cartilages, , (b) utilizing datasets corresponding to both sustained vowel phonation and connected speech, and (c) using datasets from both normophonic and disordered subjects.
This will enable a more detailed dynamic analysis of voice production in normophonic and disordered cases and potentially provides insights regarding laryngeal muscular activity.

In this study, we present a fully convolutional neural network architecture, U-Net for the aforementioned segmentation task (\citeauthor{m4}, \citeyear{m4}). U-Net is widely used for medical image segmentation due to its precise localization, strong contextual representation, and good performance with small datasets (\citeauthor{i73}, \citeyear{i73}). Different laryngeal regions were manually annotated in HSV data of normophonic and disordered voices to train, validate, and test the network. The performance of the network was assessed both visually and quantitatively using Intersection over Union (IoU), Dice score, precision, and recall. This approach has the potential to support analysis of laryngeal tissue motion and structural behaviors during voice production in normophonic and disordered voices.

\section{Methods} \label{methods}
\begin{figure}[h]
\begin{center}
\includegraphics[width=\columnwidth]{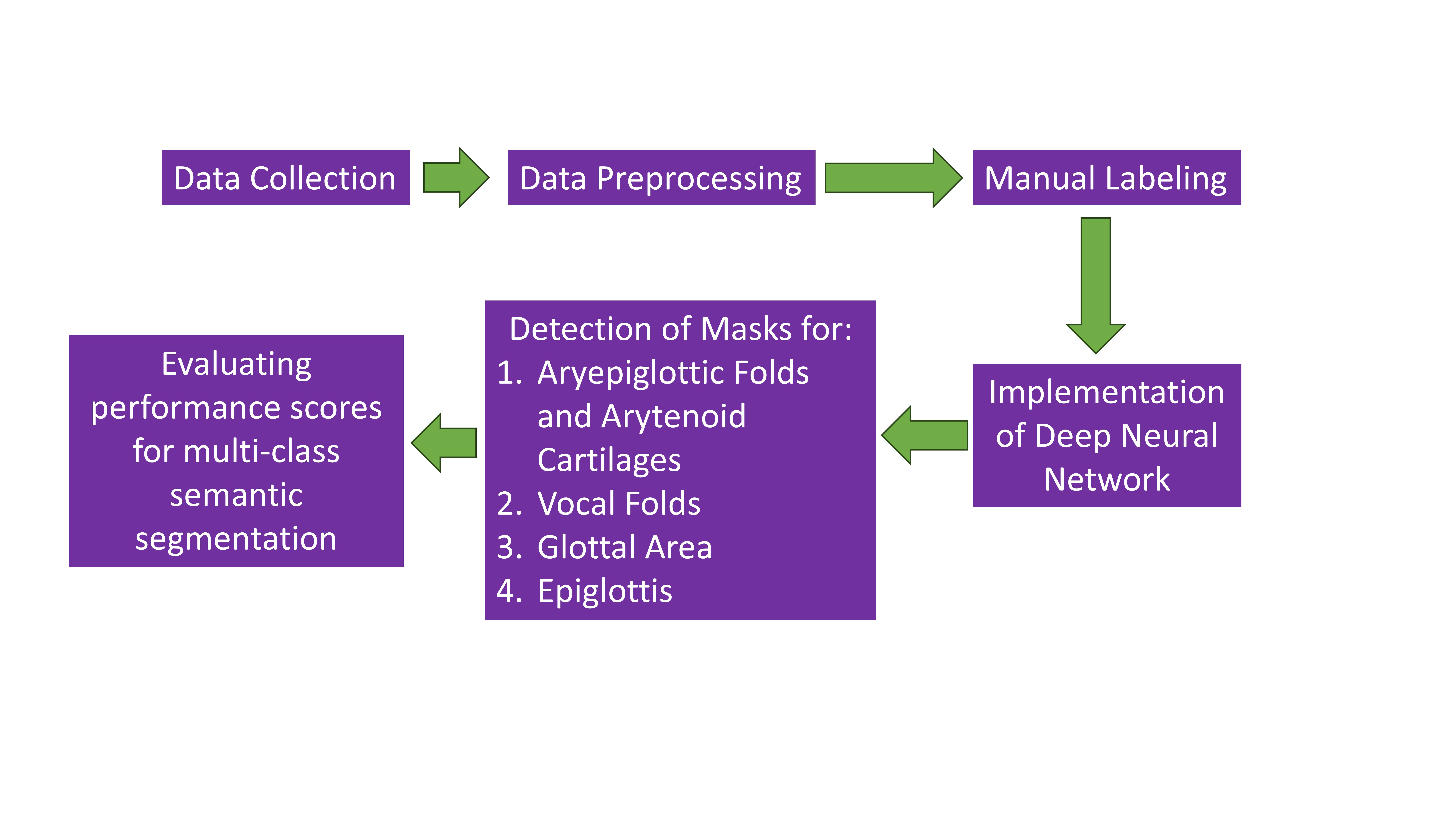}
\caption{Overview of the data collection and analysis framework for multi-class semantic segmentation of laryngeal structures.}
\label{f1}
\end{center}
\end{figure}

The data collection and analysis framework is shown in Fig. \ref{f1}. Data collection and extraction are described in Section \ref{data}, followed by data preprocessing in Section \ref{preprocess}. Manual labeling of laryngeal tissue masks is discussed in Section \ref{label}. The deep neural network implementation and training are presented in Section \ref{network}. Then, performance assessment metrics are discussed in Section \ref{performance}. Finally, training performance, mask prediction, and segmentation performance evaluation are presented in the Results section (Section \ref{results}), followed by a discussion of the findings and their implications in Section \ref{discussion} and the main conclusions in Section \ref{conclusions}.

\subsection{Data Collection and Extraction } \label{data}

\underline{\textbf{Participant Demographics:}}  
Data were collected from fourteen adult participants aged between 22 and 77 years (see Table \ref{T1}). Among them, 3 were male and the remaining participants were female (M and F in the subject IDs denote male and female). Eight participants were normophonic (subject ID starting with \textbf{N}), and six had voice disorders: one with paralysis (subject IDs starting with an \textbf{P}), three with adductor laryngeal dystonia (subject IDs starting with an \textbf{S}), and two with essential vocal tremor (subject IDs starting with a \textbf{T}).

\begin{table}[h]
\centering
\caption{Participants demographics.}
\begin{tabular}{c c c}  \label{T1}
\\ \hline
Subject ID & Sex & Age\\ \hline
N9M & Male & 49\\
N11F & Female & 52\\
N13F & Female & 35\\
N23F & Female & 29\\
N25F & Female & 24\\
N26F & Female & 22\\
N31M & Male & 22\\
N32F & Female & 46\\
P030F & Female & 68\\
S006F & Female & 67\\
S007F & Female & 60\\
S008F & Female & 76\\
T012F & Female & 77\\
T036M & Male & 65\\ \hline
\end{tabular}
\end{table}

\underline{\textbf{Data Collection:}}  
HSV data were obtained from the participants during production of sustained vowel /i/, Consensus Auditory-Perceptual Evaluation of Voice (CAPE-V) sentences, and part of the Rainbow Passage. The recordings were made using Photron FASTCAM Mini AX200 monochrome high-speed camera (Photron Inc., San Diego, CA), coupled with a flexible nasolaryngoscope.
The data were obtained with a spatial resolution of $256 \times 224$ pixels, a frame rate of 4,000 frames per second (fps), and a bit depth of 12.

\underline{\textbf{Data Extraction:}}  
The HSV recordings consisted of sequences of gray-scale images, with each pixel represented using 12 bits for its intensity value. These pixels were arranged into a $256 \times 224$ two-dimensional matrix to form each grayscale frame. Each frame was then resized to $256 \times 256$ pixels and converted to an 8-bit depth.
The image resolution was changed to allow the use of a deeper U-Net architecture. In U-Net, the spatial dimensions of the images decrease by a factor of 2 at each encoder layer (see Section \ref{network} for more details). Therefore, it is preferable to use image dimension that can be expressed as large powers of 2, allowing the dimensions to be repeatedly divided by 2.

Although the recordings were initially captured at a 12-bit depth, the effective dynamic range was limited, as the pixel values within the frames occupied only a small portion of the available intensity range. Therefore, the higher bit depth only increased the data size without providing additional image details. Converting the images to an 8-bit depth preserved the effective dynamic range, as the distribution of the pixel values occupied almost the same relative intensity range as in the 12-bit images. This conversion also reduced the data size, enabling faster data loading and processing in software and machine-learning algorithms.

\subsection{ Data Preprocessing } \label{preprocess}
Image preprocessing was applied to enhance the visualization of laryngeal structures, facilitate tissue annotation, and improve boundary definition before further analysis. The preprocessing steps included noise removal and histogram equalization.

\subsubsection{Noise Removal}
The original HSV images contained honeycomb artifact due to the nature of fiberendoscopic imaging. This noise makes image segmentation, object detection, and feature extraction more challenging. It also makes manual segmentation harder by reducing image clarity. The honeycomb noise appears as patterns of light and dark squares with intensity variations over small spatial areas, indicating that the noise is associated with high-frequency image components.
To remove the noise while keeping important image details, a low-pass filter was used to remove the high frequency noise. First, the frequency content of the image was analyzed using the discrete cosine transform (DCT). For an $M \times N$ image matrix $A(x,y)$, the two-dimensional DCT is defined as (\citeauthor{m1}, \citeyear{m1}; \citeauthor{m2}, \citeyear{m2}):

\begin{equation}
B(u,v) = \alpha(u) \alpha(v) \sum_{x=0}^{M-1} \sum_{y=0}^{N-1} A_{x,y} \cos \left[ \frac{\pi (2x+1) u}{2M} \right] \cos \left[ \frac{\pi (2y+1) v}{2N} \right]
\end{equation}
Here,
\begin{itemize}
  \item $x$ and $y$ are the pixel indices in the spatial domain.
  \item $u$ and $v$ are the frequency indices in the horizontal and vertical directions, respectively.
  \item $u$ and $v$ range from $0$ to $M-1$ and $0$ to $N-1$, respectively.
  \item $B(u,v)$ is the DCT coefficient for the frequency component $(u, v)$.
  \item $A_{x,y}$ is the pixel value at position $(x, y)$.
  \item $\alpha(u)$ and $\alpha(v)$ are normalization factors defined as follows:
\end{itemize}

\begin{equation}
\alpha(u) = 
\begin{cases} 
\sqrt{\frac{1}{M}} & \text{if } u = 0 \\
\sqrt{\frac{2}{M}} & \text{if } u > 0 
\end{cases}
\end{equation}

\begin{equation}
\alpha(v) = 
\begin{cases} 
\sqrt{\frac{1}{N}} & \text{if } v = 0 \\
\sqrt{\frac{2}{N}} & \text{if } v > 0 
\end{cases}
\end{equation}

After decomposing the image into horizontal and vertical frequency components, a low-pass filter $f(u,v)$ was applied in the frequency domain using the following equations:

\begin{equation}
f(u,v) = 
\begin{cases} 
1 & \text{if } \sqrt{u^2+v^2} \leq R \\
e^{ \frac{-{(\sqrt{u^2+v^2}-R)}^2}{2 \sigma ^2}} & \text{if } \sqrt{u^2+v^2} > R \\
\end{cases}
\end{equation}

\begin{equation}
\overline{B}(u,v)=B(u,v).*f(u,v)
\end{equation}

Here,
\begin{itemize}
  \item $f(u,v)$ is the filter applied in the frequency domain.
  \item  $\overline{B}(u,v)$ is the modified frequency response after applying the filter. This is obtained by performing element-wise multiplication of the original frequency response $B(u,v)$ with the filter $f(u,v)$.
  \item $R$ is the threshold frequency. If the Euclidean distance from the origin $(0,0)$ is below $R$, the frequency response remains unchanged (multiplied by 1). If the distance is above $R$, the response is reduced by a factor less than 1. The reduction follows a Gaussian pattern as the distance from $R$ increases. The parameter $\sigma$ controls this reduction, where larger $\sigma$ values result in less attenuation.
\end{itemize}

To reconstruct the image from the modified frequency response, the inverse 2D discrete cosine transform (IDCT) was applied. The IDCT converts the modified frequency components back into the spatial domain, producing an image with reduced noise. The formula for the inverse 2D discrete cosine transform (IDCT) is given by (\citeauthor{m1}, \citeyear{m1}; \citeauthor{m2}, \citeyear{m2}):
\begin{equation}
\overline{A}(x,y) = \sum_{u=0}^{M-1} \sum_{v=0}^{N-1} \alpha(u) \alpha(v) \overline{B}(u,v) \cos \left[ \frac{\pi (2x+1) u}{2M} \right] \cos \left[ \frac{\pi (2y+1) v}{2N} \right]
\end{equation}

Here,
\begin{itemize}
    \item $\overline{A}(x,y)$ is the reconstructed pixel value at position $(x, y)$,
    \item $\alpha(u)$ and $\alpha(v)$ are the same normalization factors as defined earlier.
\end{itemize}

\subsubsection{Histogram Equalization}
The pixel intensities of the original HSV images were distributed within a narrow range, resulting in low contrast that could obscure tissue boundaries. To improve contrast, histogram equalization was applied to redistribute the pixel intensities over a wider range and enhance the visibility of laryngeal structures \citep{i64}. This process uses the cumulative distribution of pixel intensities to map the original intensity values to a broader range. As a result, differences between regions with similar intensity values become more distinguishable.

\subsection{ Labeling of Laryngeal Tissue Masks } \label{label}
\begin{figure}[h]
\begin{center}
\includegraphics[width=\columnwidth]{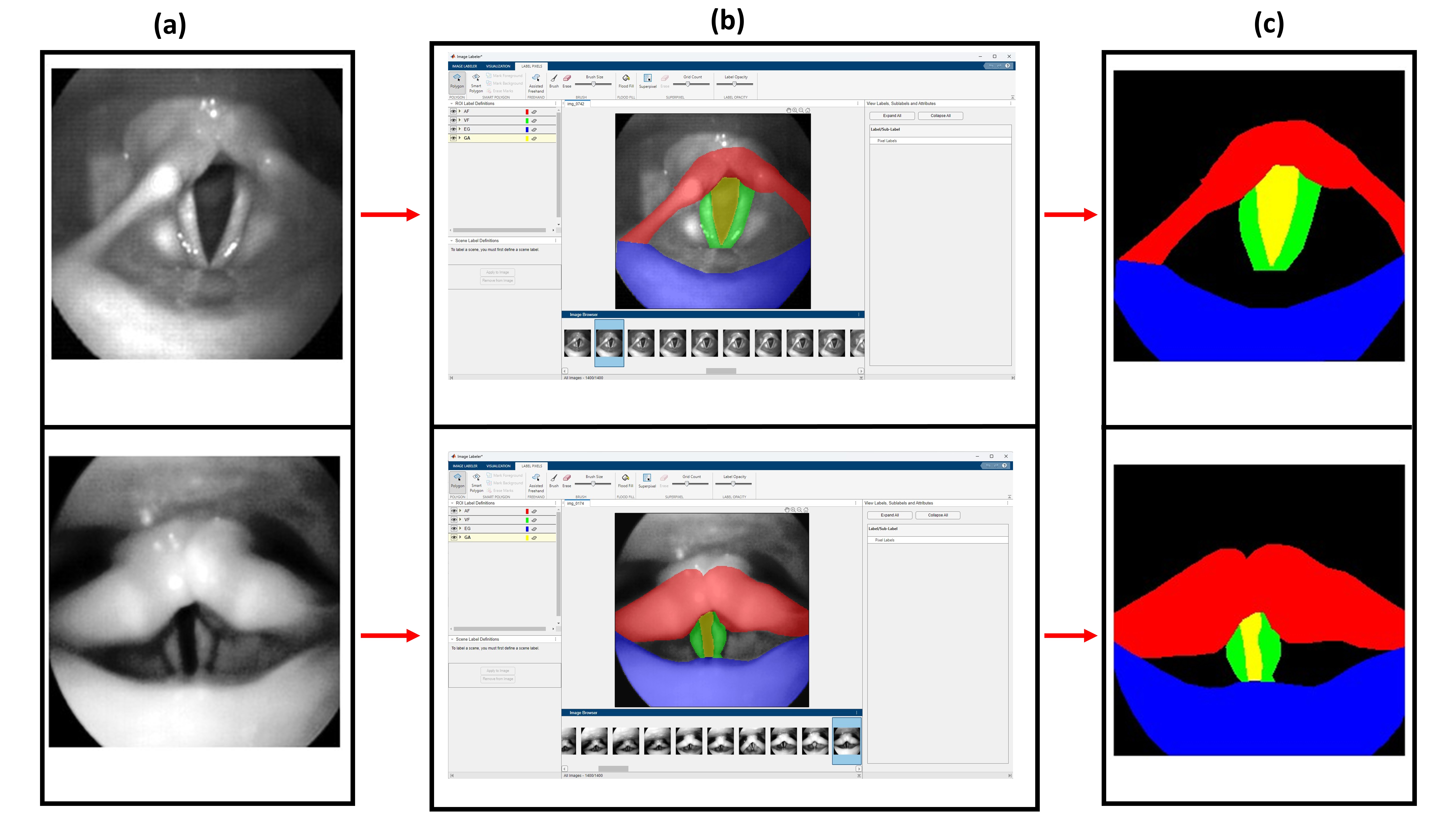}
\caption{Manual annotation of different laryngeal structures in HSV frames. (a) Preprocessed HSV frames. (b) Annotated masks overlaid on the HSV frames. (c) Corresponding segmentation masks showing the aryepiglottic folds and arytenoids (in red), vocal folds (in green), glottal area (in yellow), epiglottis (in blue), and background (in black).}
\label{f5}
\end{center}
\end{figure}

Labeling masks for different laryngeal structures is required to train the deep learning network. The aryepiglottic folds and arytenoids, vocal folds, glottal area, and epiglottis were manually annotated using the ``Image Labeler'' app in ``MATLAB R2024a'' software, as illustrated in Fig. \ref{f5}. A total of 1,400 images were labeled from 14 subjects, with 100 annotated images from each subject, covering different maneuvers and positions. The aryepiglottic folds and arytenoids were assigned a value of 1, the vocal folds a value of 2, the glottal area a value of 3, and the epiglottis a value of 4. All regions outside these structures were considered background and assigned a value of 0.

The labels were one-hot encoded, with each class represented by a separate channel. If a pixel belonged to a specific class, it took a value of 1 in that class channel and a value of 0 in the other channels. As a result, each label mask with a dimension of $256 \times 256$ (matching the grayscale image resolution) was converted to $5 \times 256 \times 256$, where 5 represents the total number of classes (0-4).

\subsection{ Implementation of Deep Neural Network} \label{network}
\subsubsection{ Deep Neural Network Architecture}
\begin{figure}[h]
\begin{center}
\includegraphics[width=\columnwidth]{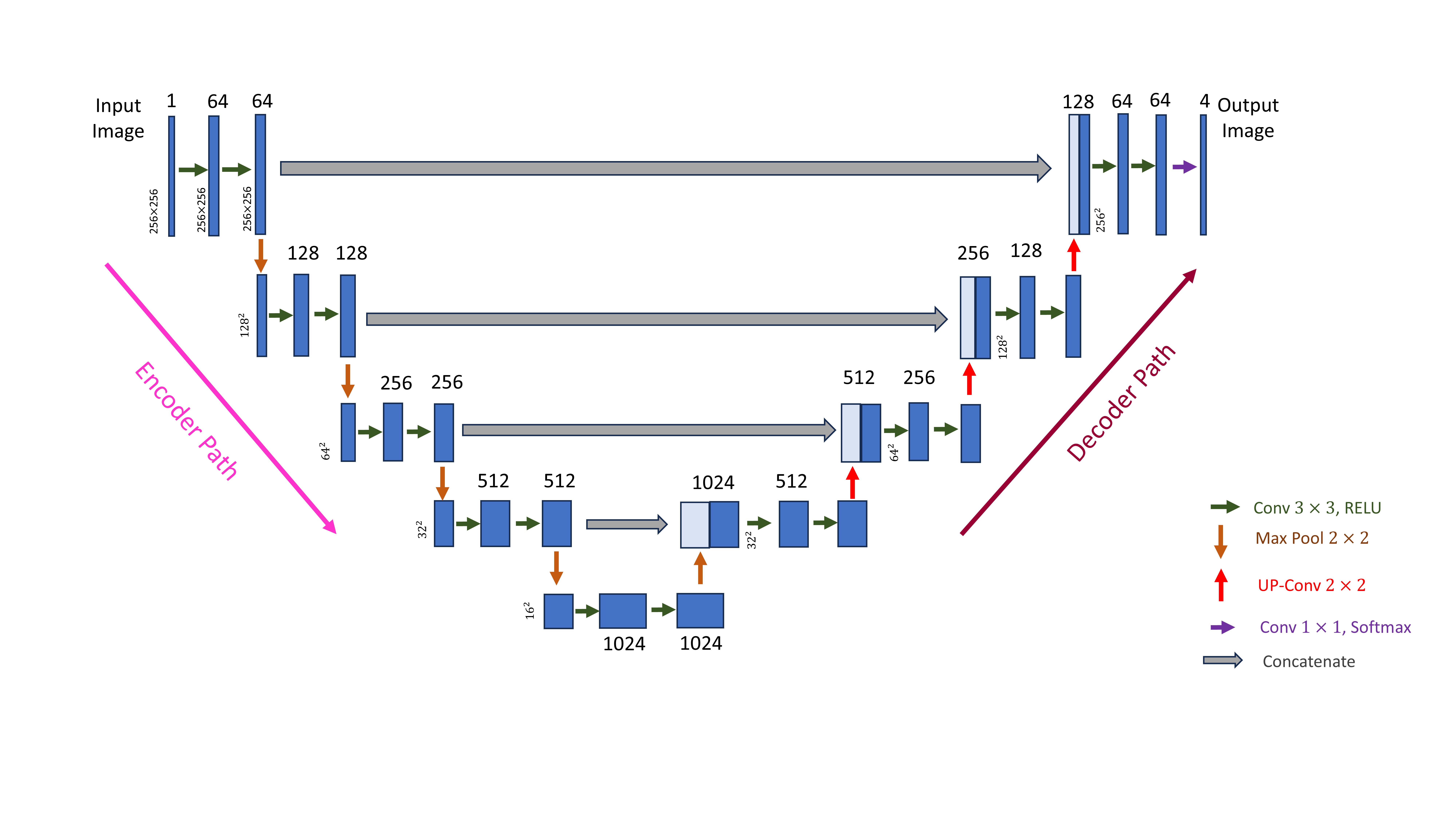}
\caption{Schematic of the U-Net architecture consisting of five encoder and decoder stages, with feature-channel depths of 64, 128, 256, 512, and 1024 in both the encoder and decoder paths. Arrows represent different operations as shown in the legend at the lower right corner.}
\label{f6}
\end{center}
\end{figure}

For network training, a U-Net architecture (\citeauthor{m4}, \citeyear{m4}) consisting entirely of convolutional layers, without any dense layers, was used. The network has two symmetric paths: an encoder and a decoder, as shown in Fig. \ref{f6}. In the encoder path, features are extracted through successive convolutional layers, while the image resolution progressively decreases and the number of feature channels increases, allowing the network to learn increasingly complex image features. In the decoder path, this process is reversed. The encoder captures contextual information, while the decoder provides precise localization for semantic segmentation.

At the beginning of each encoder layer, the number of channels is doubled.
Each encoder layer consists of two blocks, where each block contains a $3 \times 3$ convolutional layer followed by ReLU activation.
Finally, after these blocks, $2 \times 2$ max-pooling operation is applied, reducing the spatial dimensions in both the row and column directions by a factor of 2. 
The bottleneck layer connects the encoder and decoder paths. In this layer, the image has the lowest spatial resolution and the highest number of channels. This layer is similar to the encoder convolutional layers, except that no max-pooling operation is applied after the two repeated convolutional layers followed by ReLU activation.
At the beginning of each decoder layer, the image resolution is doubled using a 2D transposed convolution. Then, the image in the decoder layer is concatenated with the image from the encoder layer that has the same spatial resolution and the same number of channels, thereby doubling the number of channels. The number of channels is then halved, followed by two repeated convolutional layers with ReLU activation.

In this study, each input image had a single channel since grayscale images were used. Before entering the first encoder layer, the number of image channels was matched to the number of channels in the first encoder layer. At the output, the number of channels was five, corresponding to masks for five different zones (including the background). Therefore, at the output, the number of image channels was changed from the number of channels in the last decoder layer to five.

U-Net can have different numbers of encoder and decoder stages, but the number of stages in the encoder and decoder must be the same. In this study, three U-Net configurations were implemented: a 4-layer network with 64–128–256–512 channels (denoted as UNET-4), a 5-layer network with 64–128–256–512–1024 channels (denoted as UNET-5), and a 6-layer network with 64–128–256–512–1024–2048 channels (denoted as UNET-6).
Batch normalization was applied after each convolution operation to enable faster and more stable training and to prevent gradient vanishing or explosion. To reduce overfitting, dropout was used and 5\% of the total weights were randomly disabled during training.

\subsubsection{Dataset Splitting and Network Training}
The HSV 1,400 images and their corresponding annotated labels were divided into training, validation, and test sets. After splitting the dataset, the training set contained 1,050 images (75\% of the total), the test set contained 280 (20\% of the total), and the validation set contained 70 images (5\% of the total).
The number of images in the training set was increased to provide more diverse training instances, allowing the model to perform better on different types of images. Hence, the training data were augmented using random rotation, translation, and scaling, resulting in 5,250 images in the training set.

The training process was divided into multiple epochs. In each epoch, all training data were used to update the neural network weights through backpropagation. Loss minimization (formula for loss function used in the model is provided in Eq. \ref{loss}) was performed using the Adam optimizer, which provides fast convergence and stable training by adaptively adjusting learning rates (\citeauthor{m5}, \citeyear{m5}). The weights were updated using the computed gradients multiplied by the learning rate, which was initially set to 0.0001. Each epoch was further divided into several iterations so that all training data were not processed at the same time, as this would exceed memory limits. Instead, each iteration used a smaller portion of the training data. The number of images and labels used in each iteration is called the minibatch size, which was set to 8. During the iterations of each epoch, different subsets of the training data were used so that, after all iterations were completed, the entire training dataset had been used for training. 
After each epoch, model performance was evaluated on both the training and validation datasets by calculating accuracy and loss to monitor overfitting or underfitting.
Once all epochs were completed, the training process ended. After completing training, model performance was evaluated on the test set.

\begin{figure}[h]
\begin{center}
\includegraphics[width=\columnwidth]{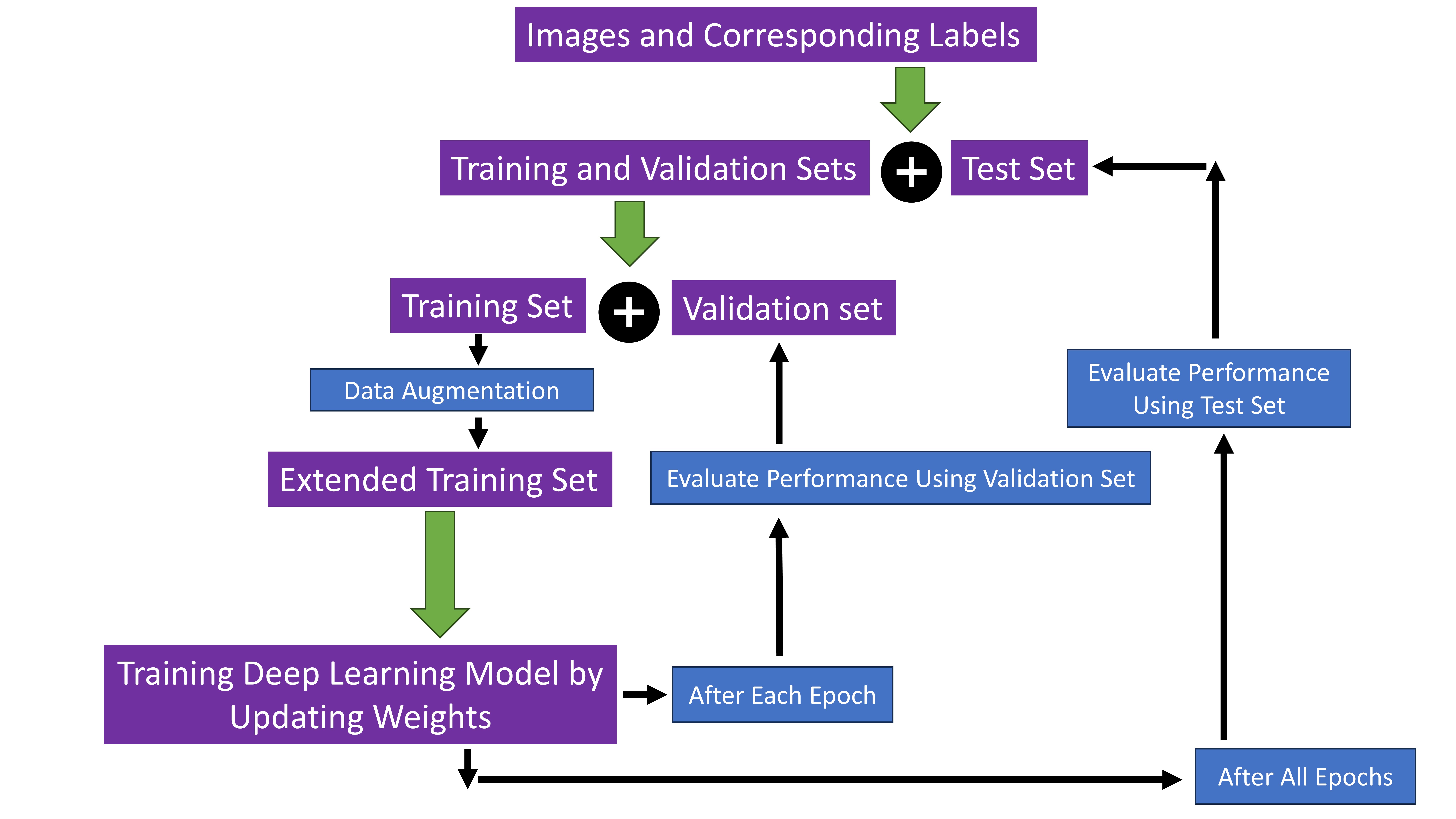}
\caption{Deep learning workflow showing dataset splitting, data augmentation, model training, and performance evaluation using training, validation, and test sets.}
\label{f7}
\end{center}
\end{figure}

All steps and sequences, from dataset splitting to model performance evaluation, are shown in Fig. \ref{f7}.

\subsection{ Performance Assessment Metrics } \label{performance}
The goal of training the deep neural network was to reduce the loss, or the difference between the ground-truth (manually annotated) labels and the labels predicted by the network. The mean categorical cross-entropy loss function (\citeauthor{m6}, \citeyear{m6}) was used, which is defined as:

\begin{equation}\label{loss}
\ell = -\frac{1}{N}\sum_{i=1}^{N}\sum_{c=1}^{C} y_{i,c}\log\left(\frac{\exp(\hat{y}_{i,c})}{\sum_{k=1}^{C}\exp(\hat{y}_{i,k})}\right)
\end{equation}

Here,
\begin{itemize}
  \item $\ell$ denotes the mean multi-class categorical cross-entropy loss over a mini-batch.
  \item $N$ is the total number of samples (or pixels) in the mini-batch.
  \item $C$ is the number of classes.
  \item $y_{i,c}$ is the ground-truth label for class $c$ of the $i$-th sample, represented using one-hot encoding.
  \item $\hat{y}_{i,c}$ is the model output for class $c$ of the $i$-th sample.
  \item $\exp(\cdot)$ denotes the exponential function.
  \item $\log(\cdot)$ denotes the natural logarithm.
\end{itemize}

The predicted probabilities for different class labels produced by the machine learning model were obtained using the softmax function (the expression inside the parentheses in Eq. \ref{loss}), which converted the model outputs into class probabilities. For the $i$-th sample (or pixel) and class $c$, the predicted probability is given by:

\begin{equation} \label{softmax}
z_{i,c} = \frac{\exp(\hat{y}_{i,c})}{\sum_{k=1}^{C} \exp(\hat{y}_{i,k})}
\end{equation}

Referring to Eq. \ref{softmax}, the softmax output $z_{i,c}$ represents a probability value between $0$ and $1$, and the probabilities across all classes sum to one.

To obtain discrete predicted labels, the softmax probabilities were converted into binary values, similar to the ground-truth labels. The class with the maximum probability was assigned a value of $1$, while all other classes were assigned a value of $0$. The resulting hard predicted label is defined as:

\begin{equation}
Z_{i,c} = 
\begin{cases}
1, & \text{if } c = \arg\max\limits_{k} z_{i,k}, \\
0, & \text{otherwise}.
\end{cases}
\end{equation}

The average training accuracy per iteration is defined as:

\begin{equation}\label{eqacc}
\mathrm{Accuracy}
=
\frac{1}{N}
\sum_{i=1}^{N}
\mathbf{1}
\left(
\arg\max_{c} Z_{i,c}
=
\arg\max_{c} y_{i,c}
\right)
=
\frac{TP + TN}
{TP + TN + FP + FN}
\end{equation}

When both the predicted label and the ground truth are 0, it is called a True Negative (TN). When the predicted label is 1 and the ground truth is 0, it is called a False Positive (FP). When the predicted label is 0 and the ground truth is 1, it is called a False Negative (FN). 
The average accuracy can also be expressed in terms of these quantities, as shown on the right-hand side of Eq. (\ref{eqacc}).
When the predicted label and the ground-truth label at a given spatial pixel location match, the accuracy is increased by one; otherwise, it remains unchanged. When both the predicted label and the ground truth are 1, it is called a True Positive (TP). 
Using the formulas in Eqs.~\ref{loss} and \ref{eqacc}, the average loss and average accuracy were calculated for the training and validation sets, after each epoch.

To evaluate the performance of the trained model on the test set, two primary metrics were used: Intersection over Union (IoU) and the Dice score (also known as the F1 score). IoU is defined as the ratio of the area of intersection to the area of union between the predicted and ground-truth labels. It is expressed by:
\begin{equation}
\mathrm{IoU}
=
\frac{\text{Intersection}}{\text{Union}}
=
\frac{TP}{TP + FP + FN}
\end{equation}

The Dice score measures the similarity between the predicted segmentation and the ground-truth segmentation. The Dice score is calculated using the following equation:
\begin{equation}
\mathrm{Dice}
=
\frac{2\,\text{Intersection}}
{\text{Prediction} + \text{Ground Truth}}
=
\frac{2TP}{2TP + FP + FN}
\end{equation}

In addition, a confusion matrix was computed for each model to provide a detailed assessment of the model’s classification behavior on the test images. 
In the confusion matrix, rows represent the actual (ground truth) classes and columns represent the predicted classes. The diagonal elements show the correctly classified samples, or true positives for each class. The off-diagonal elements represent errors. For a given class, the values in the same row but in different columns are False Negatives (FN), meaning samples that truly belong to that class but are predicted as another class. Similarly, the values in the same column but in different rows are False Positives (FP), meaning samples from other classes that are incorrectly predicted as that class.
The accuracy on the test set can be calculated from the confusion matrix by dividing the sum of the diagonal elements by the total number of elements in the matrix, which is equivalent to Eq. \ref{eqacc}.
Moreover, per-class accuracy (recall) and per-class precision can be calculated from the confusion matrix as follows:

\begin{equation}\label{eqrec}
\text{Recall}_i =
\frac{\text{Diagonal element of class } i}
{\text{Sum of row } i}
=
\frac{TP_i}{TP_i + FN_i}
\end{equation}

\begin{equation}\label{eqprec}
\text{Precision}_i =
\frac{\text{Diagonal element of class } i}
{\text{Sum of column } i}
=
\frac{TP_i}{TP_i + FP_i}
\end{equation}
In Eqs. \ref{eqrec} and \ref{eqprec}, the subscript $i$ denotes the corresponding class (0-4).

\subsubsection{Hardware and Software Configuration}
The model was executed on a system with the configuration listed in Table \ref{T2}.

\begin{table}[h]
\centering
\caption{System configuration used for model training and evaluation.}
\begin{tabular}{c c}  \label{T2}
\\ \hline
Component & Specification \\ \hline
Operating System & Windows 11 Pro (64-bit) \\
Processor & AMD Ryzen 9 9900X (12-core, 24-thread) \\
RAM & 64 GB \\
GPU & NVIDIA GeForce RTX 5080 \\
GPU VRAM & 16 GB \\
Development Environment & Python 3.13.5 (Anaconda distribution) \\
ML Framework & PyTorch 2.10.0 (CUDA 12.8) \\ \hline
\end{tabular}
\end{table}

\section{ Results } \label{results}
\begin{figure}[h]
\begin{center}
\includegraphics[width=\columnwidth]{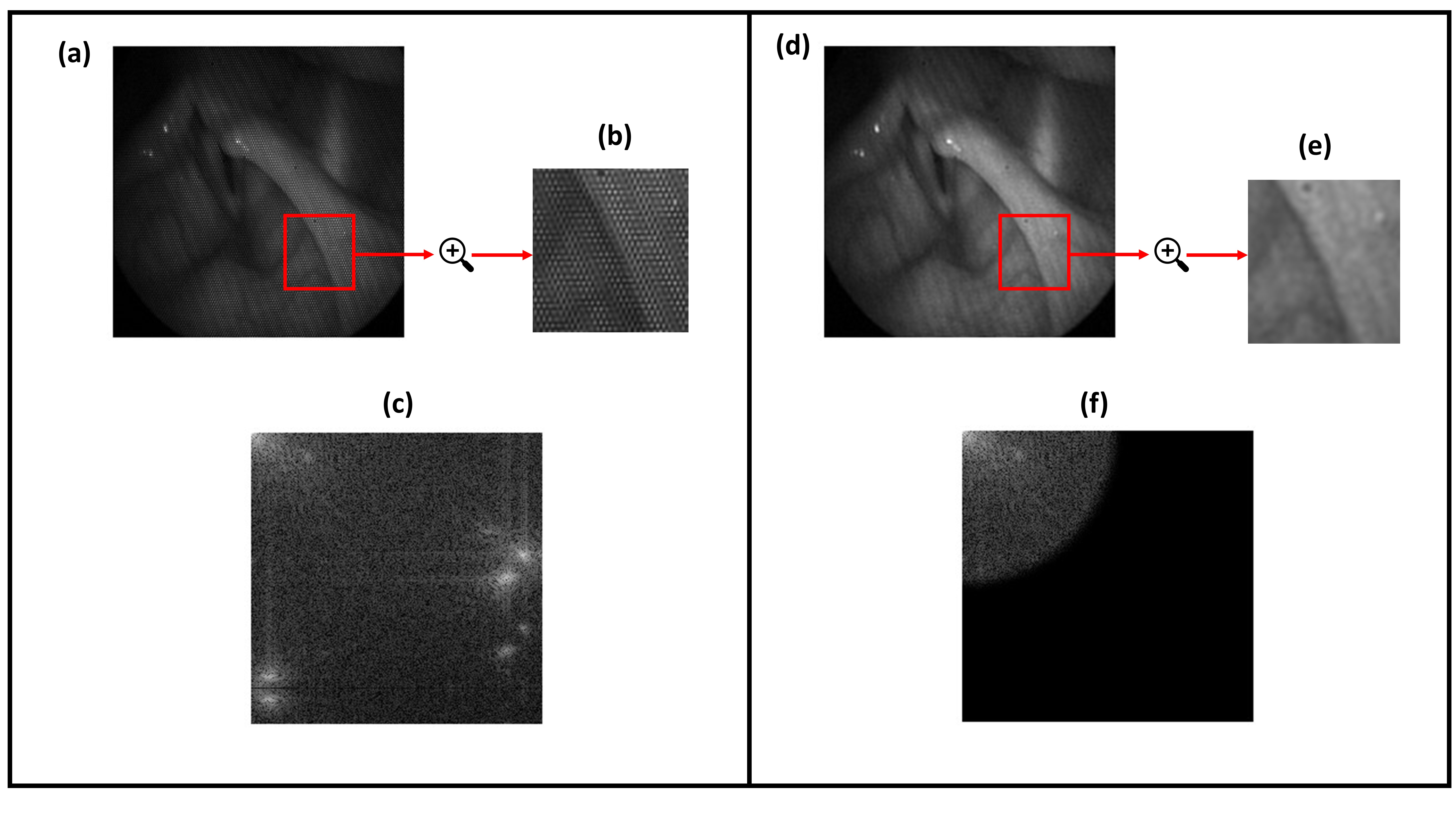}
\caption{Noise removal from an HSV frame using a low-pass filter in the frequency domain based on the discrete cosine transform (DCT). The data before and after the noise removal are shown in the left and right blocks. Panel (a) and (d) show the HSV images before and after the noise removal, respectively. Panel (b) and (e) show zoomed-in regions at the same location before and after the noise removal. Panel (c) and (f) show the corresponding DCT representations.}
\label{f3}
\end{center}
\end{figure}

Before training or testing the deep neural networks, all the utilized frames were preprocessed using noise removal followed by histogram equalization.
The effect of noise removal and the corresponding frequency-domain coefficients are shown in Fig. \ref{f3}. The left block in the figure includes the data before the noise removal while the right block includes the data after the noise removal. 
Panels (c) and (f) in Fig. \ref{f3} show the DCT coefficients before and after the noise removal, respectively. In these panels, brighter regions indicate larger DCT coefficients, while darker regions indicate smaller coefficients. Pixel positions farther from the top-left corner represent higher-frequency components. After the noise removal, the higher-frequency components beyond a certain radius from the top-left corner are attenuated, as shown in Fig. \ref{f3}f.

\begin{figure}[h]
\begin{center}
\includegraphics[width=\columnwidth]{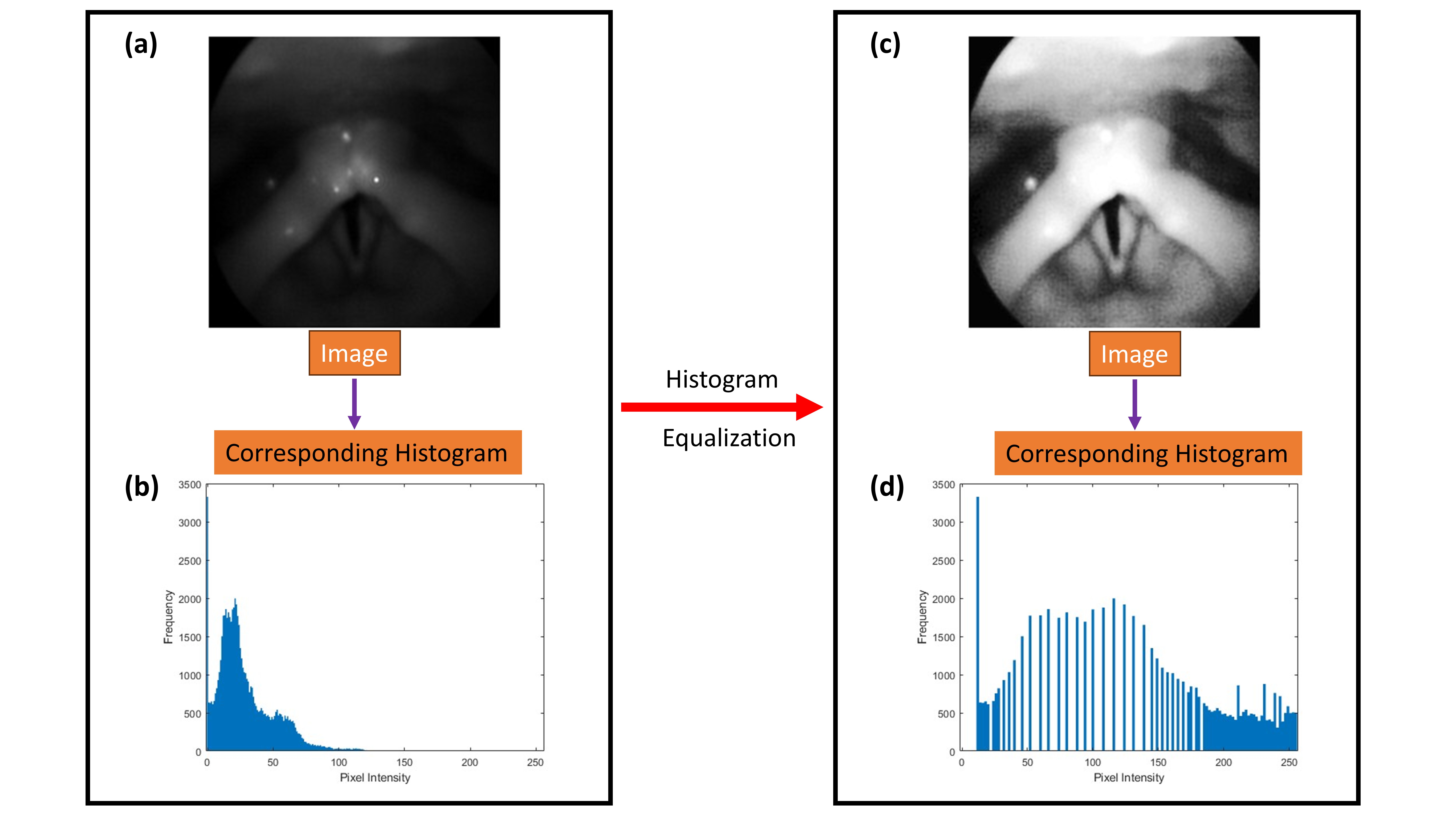}
\caption{Enhancement of image contrast using histogram equalization. Top: HSV images. Bottom: corresponding histograms. Left: before histogram equalization. Right: after histogram equalization.}
\label{f4}
\end{center}
\end{figure}

Fig. \ref{f4} shows a comparison of an HSV frame before (left panel) and after (right panel) contrast enhancement using histogram equalization.
Fig. \ref{f4}b and d represent the histograms of pixel values before and after histogram equalization, respectively. The pixel values were distributed within a small range before the histogram equalization, whereas after the histogram equalization, they were distributed across the entire intensity range (see Fig. \ref{f4}b and d).

\begin{figure}[h]
\centering

\begin{subfigure}[t]{0.32\columnwidth}
    \centering
    \caption{}
    \includegraphics[width=\columnwidth]{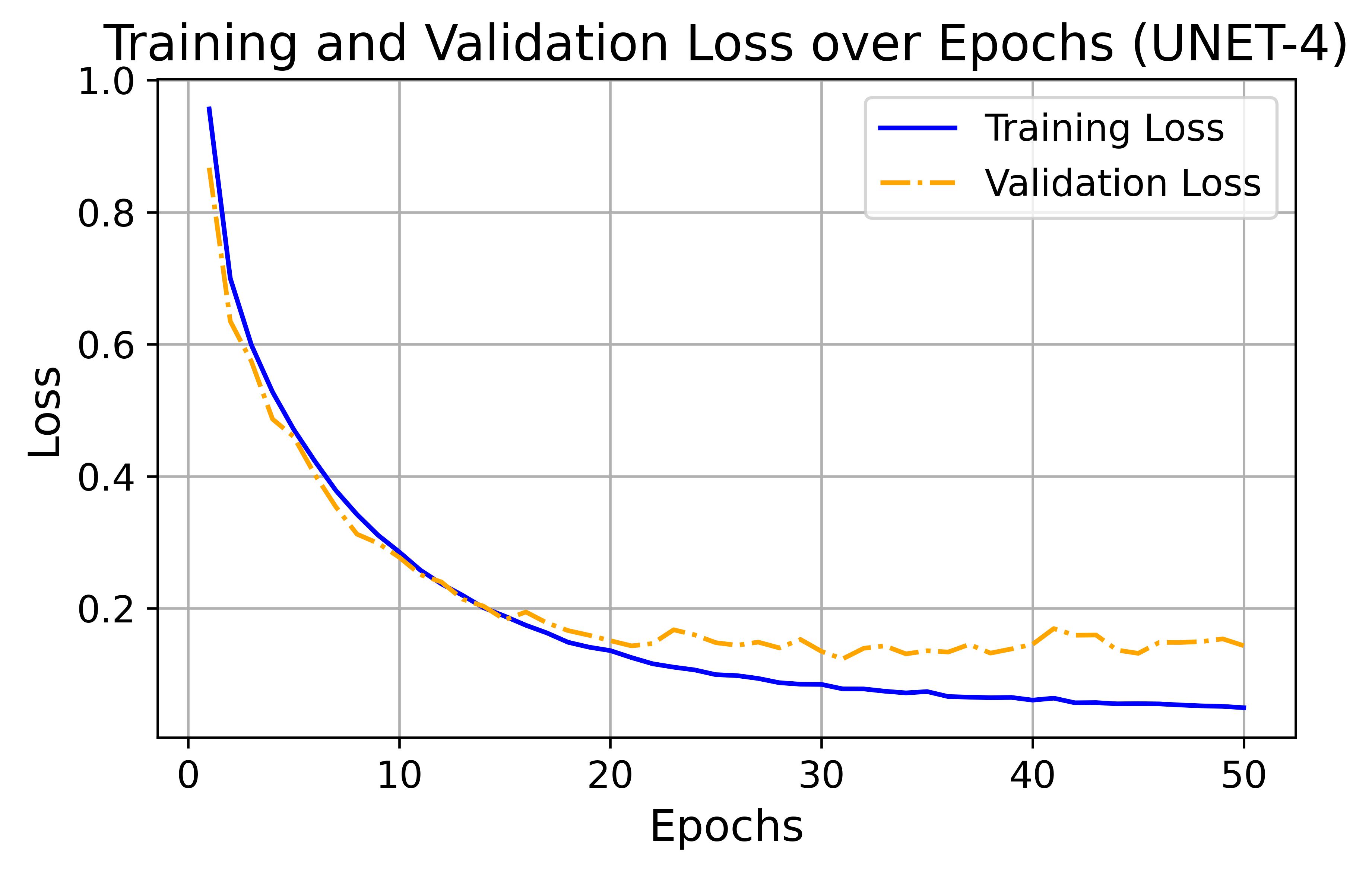}
    \label{fig:8a}
\end{subfigure}
\hfill
\begin{subfigure}[t]{0.32\columnwidth}
    \centering
    \caption{}
    \includegraphics[width=\columnwidth]{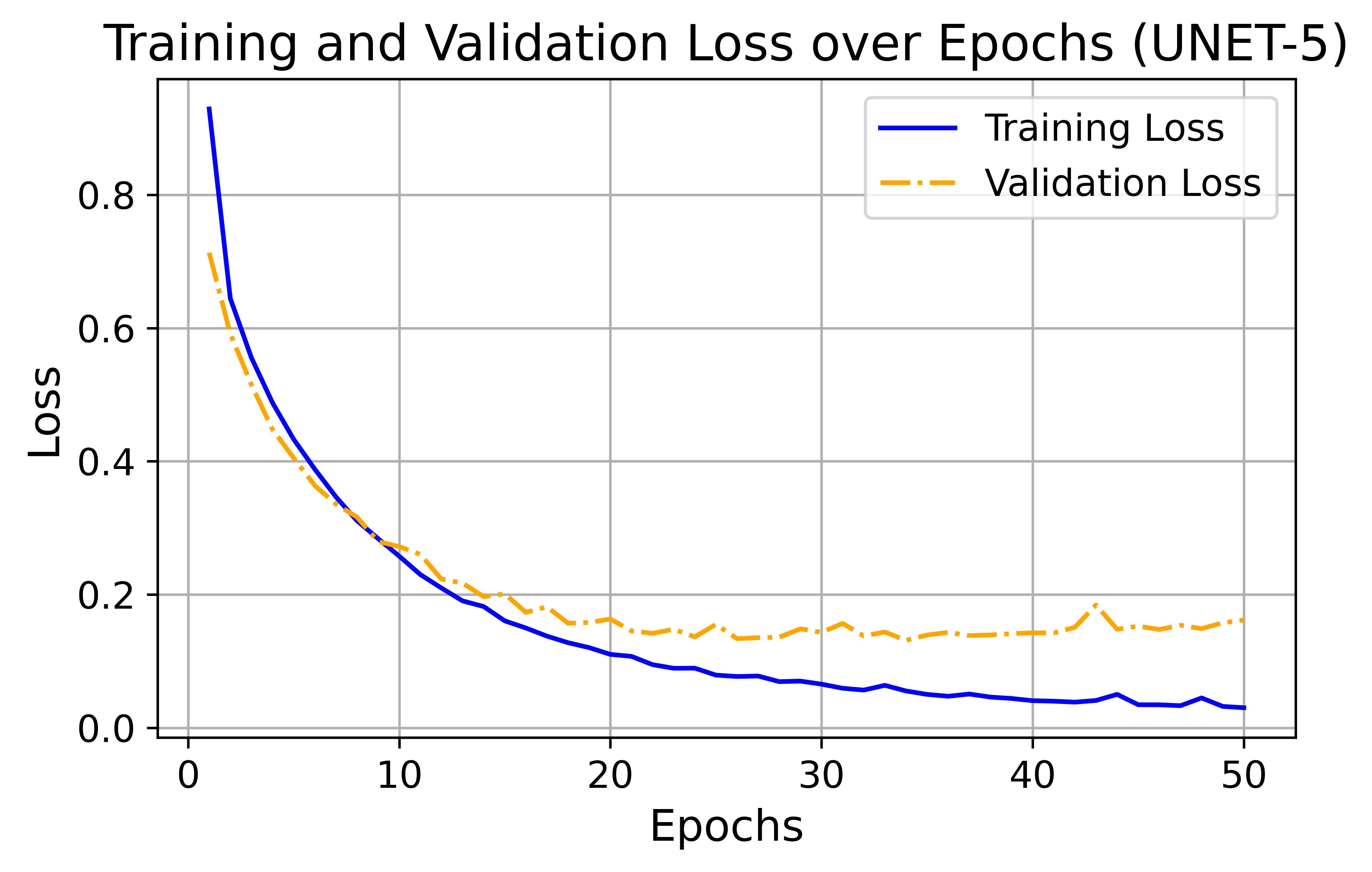}
    \label{fig:8b}
\end{subfigure}
\hfill
\begin{subfigure}[t]{0.32\columnwidth}
    \centering
    \caption{}
    \includegraphics[width=\columnwidth]{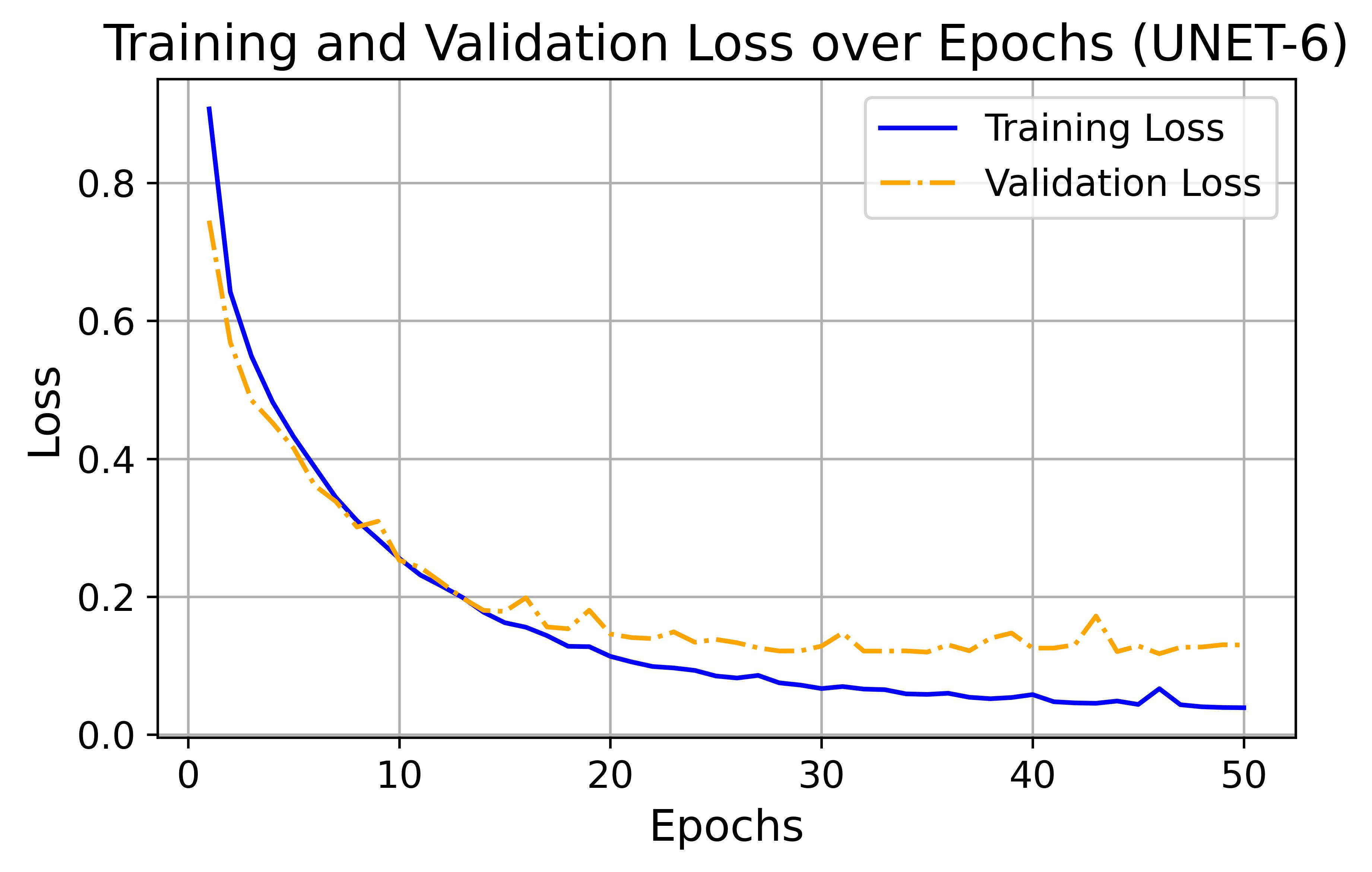}
    \label{fig:8c}
\end{subfigure}

\caption{Training and validation loss over epochs for the three U-Net models during training: (a) UNET-4, (b) UNET-5, and (c) UNET-6. Blue solid lines indicate the training loss, and orange dashed lines show the validation loss.}
\label{f8}

\end{figure}

\begin{figure}[h]
\centering

\begin{subfigure}[t]{.32\columnwidth}
    \centering
    \caption{}
    \includegraphics[width=\columnwidth]{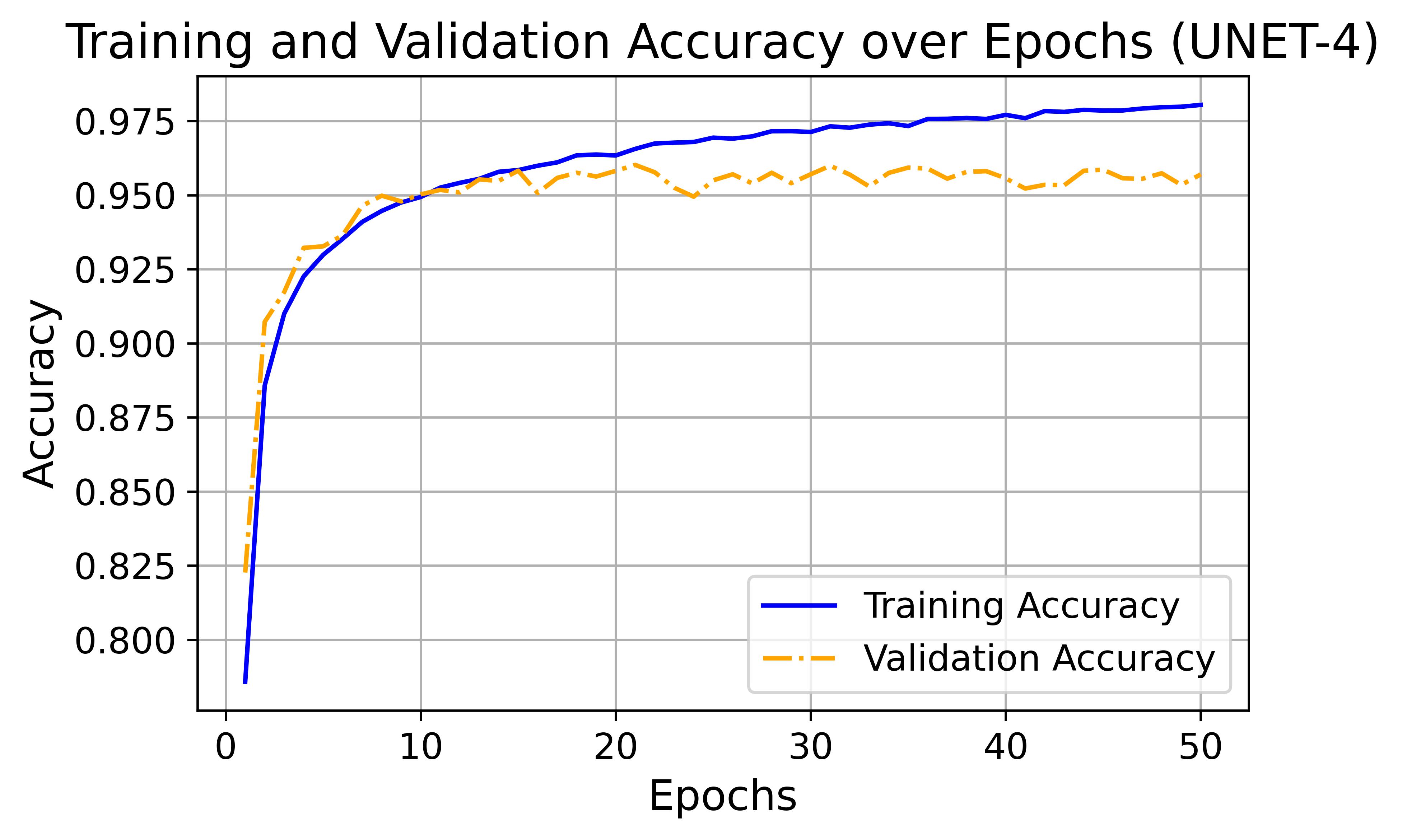}
    \label{fig:9a}
\end{subfigure}
\hfill
\begin{subfigure}[t]{0.32\columnwidth}
    \centering
    \caption{}
    \includegraphics[width=\columnwidth]{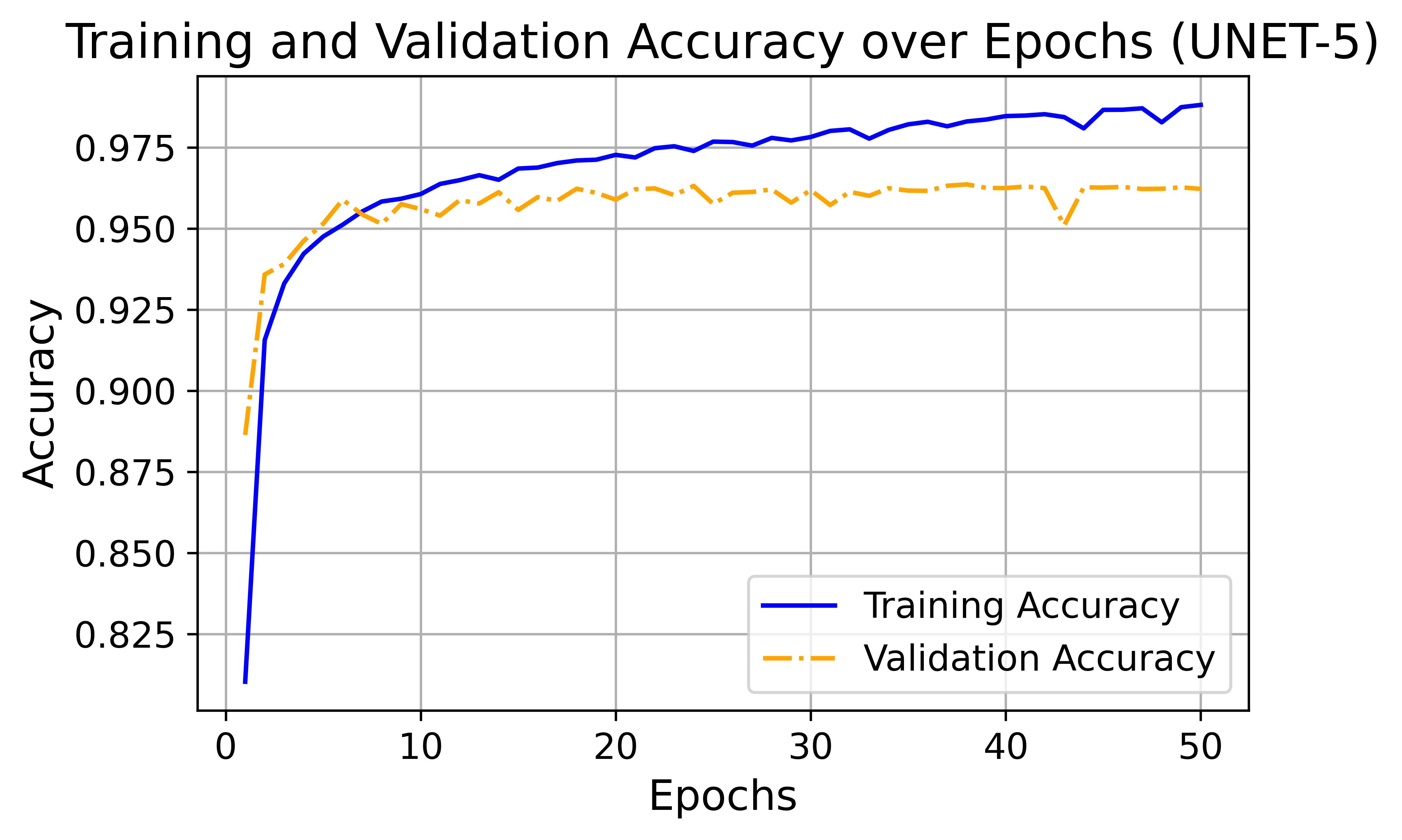}
    \label{fig:9b}
\end{subfigure}
\hfill
\begin{subfigure}[t]{0.32\columnwidth}
    \centering
    \caption{}
    \includegraphics[width=\columnwidth]{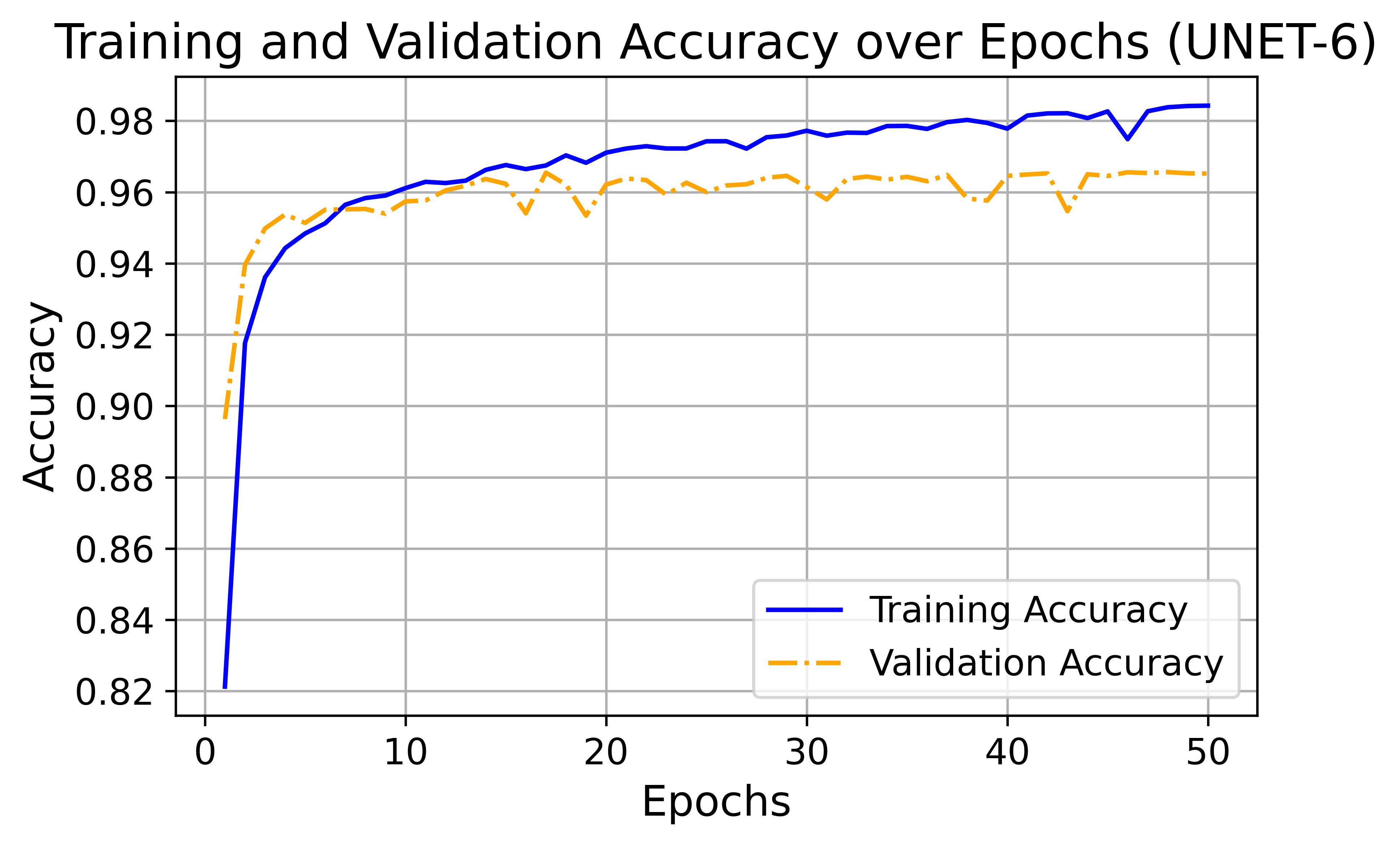}
    \label{fig:9c}
\end{subfigure}

\caption{Training and validation accuracy over epochs for the three U-Net models during training: (a) UNET-4, (b) UNET-5, and (c) UNET-6. Blue solid lines: training accuracy; Orange dashed lines: validation accuracy.}
\label{f9}

\end{figure}

Following the preprocessing steps and manual annotation, the frames in the training set were used to train three U-Net models.
Each of the three models was trained for 50 epochs. After each epoch, the training and validation loss and accuracy were computed to monitor the learning progress. 
Although slight fluctuations are observed in the training and validation loss and accuracy, the overall trend for all three models shows a decrease in training and validation loss over the epochs, as illustrated in Fig.~\ref{f8}. In contrast, the general trend of training and validation accuracy for all models increases progressively with each epoch (see Fig.~\ref{f9}).

For all models, the lowest training loss and highest training accuracy occur at the same epoch, which is the final epoch.
This behavior is expected because, during training, the model parameters are updated in each epoch based on the training data to reduce the loss and improve the match between the predicted outputs and the ground truth labels.
On the other hand, the maximum validation accuracy and the minimum validation loss do not occur at the same epoch, since the models do not use the validation data to update their parameters. 
For UNET-4, UNET-5, and UNET-6, the minimum validation loss occurs at epochs 30, 33, and 45, respectively, while the maximum validation accuracy is achieved at epochs 20, 37, and 47.
After completing all epochs for each model, the final model weights were selected based on the lowest validation loss in order to prevent overfitting. 
Subsequently, the trained models were applied to the test set to predict the labels and compute different performance metrics.

\begin{figure}[h] 
\begin{center}
\hspace*{-.45cm}
\includegraphics[height=0.4\textheight]{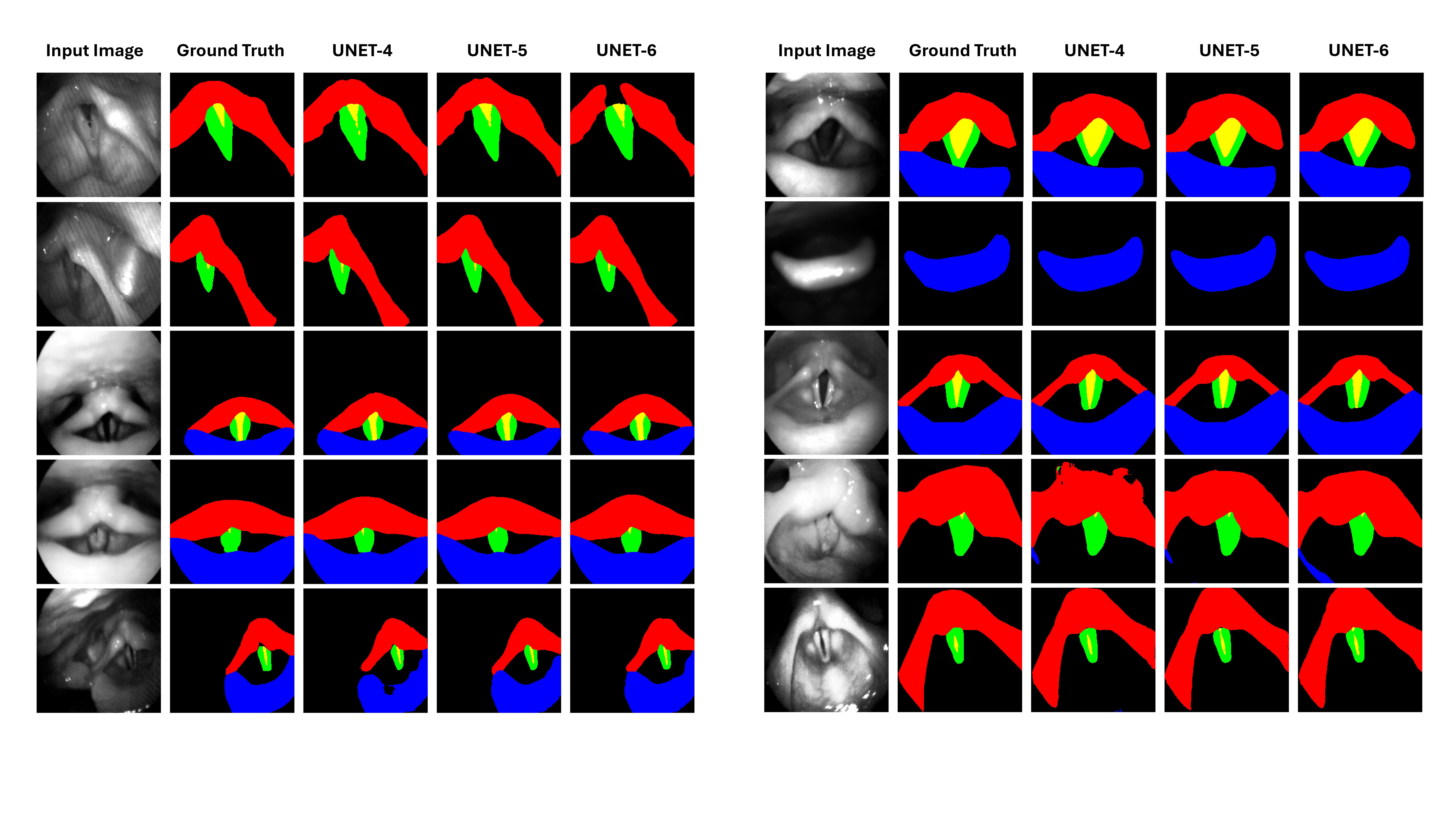}
\caption{Comparison of segmentation results predicted by three U-Net models on ten representative test images, arranged in two panels. In each panel, the images are organized row-wise. From left to right: input image, ground truth labels, and predictions from UNET-4, UNET-5, and UNET-6. Black: background; red: aryepiglottic folds and arytenoids; green: vocal folds; blue: epiglottis; yellow: glottal area.}
\label{f10}
\end{center}
\end{figure}

The predicted labels are compared with the ground truth labels as shown in Fig.~\ref{f10}. 
Due to the smaller number of encoder–decoder layers, UNET-4 reduces the spatial resolution much less than UNET-5 and UNET-6. As a result, using the same $3 \times 3$ convolutional layers, it cannot capture long-range spatial relationships effectively.
 In contrast, UNET-6, which has more encoder–decoder layers, reduces the spatial resolution further and is therefore able to capture more distant spatial relationships.
However, due to the larger number of trainable parameters in UNET-6, it has a higher tendency to overfit. For example, as shown in the fourth image from the top in the second panel of Fig.~\ref{f10}, UNET-6 predicts the largest epiglottis region even though the image does not contain the epiglottis.
Based on visual inspection, UNET-6 shows the best overall performance, while UNET-5 performs almost similarly to UNET-6.
In contrast, UNET-4 is less reliable in some cases, and the predicted label boundaries are not as smooth as those produced by UNET-5 and UNET-6.
However, in most cases, all three networks reliably capture the glottal area, which is the most important region for many studies and analyses.
The networks are able to accurately predict different degrees of glottal opening.

\begin{figure}[h]
\centering

\begin{subfigure}[t]{0.32\columnwidth}
    \centering
    \caption{}
    \includegraphics[width=\columnwidth]{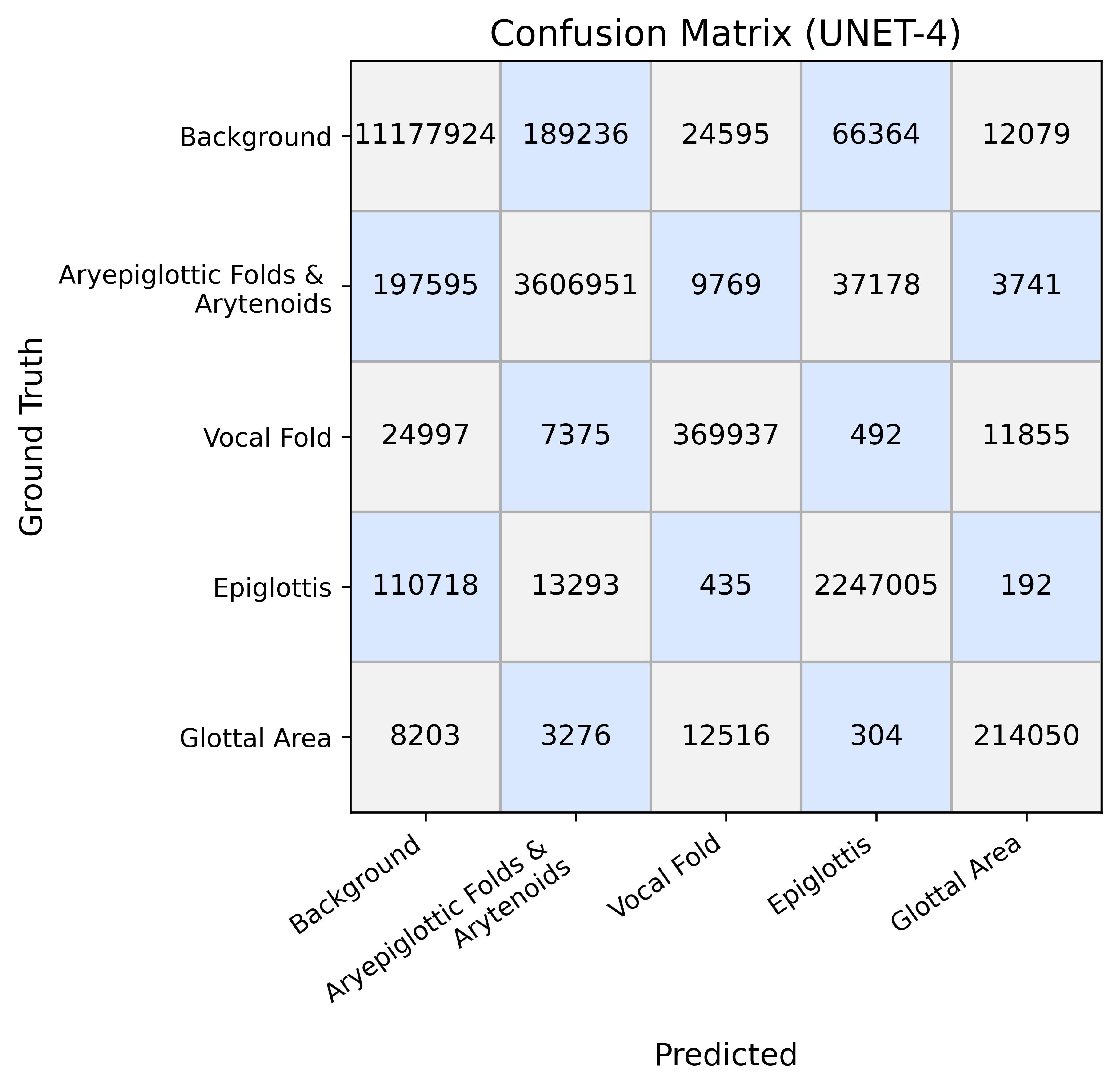}
    \label{fig:11a}
\end{subfigure}
\hfill
\begin{subfigure}[t]{0.32\columnwidth}
    \centering
    \caption{}
    \includegraphics[width=\columnwidth]{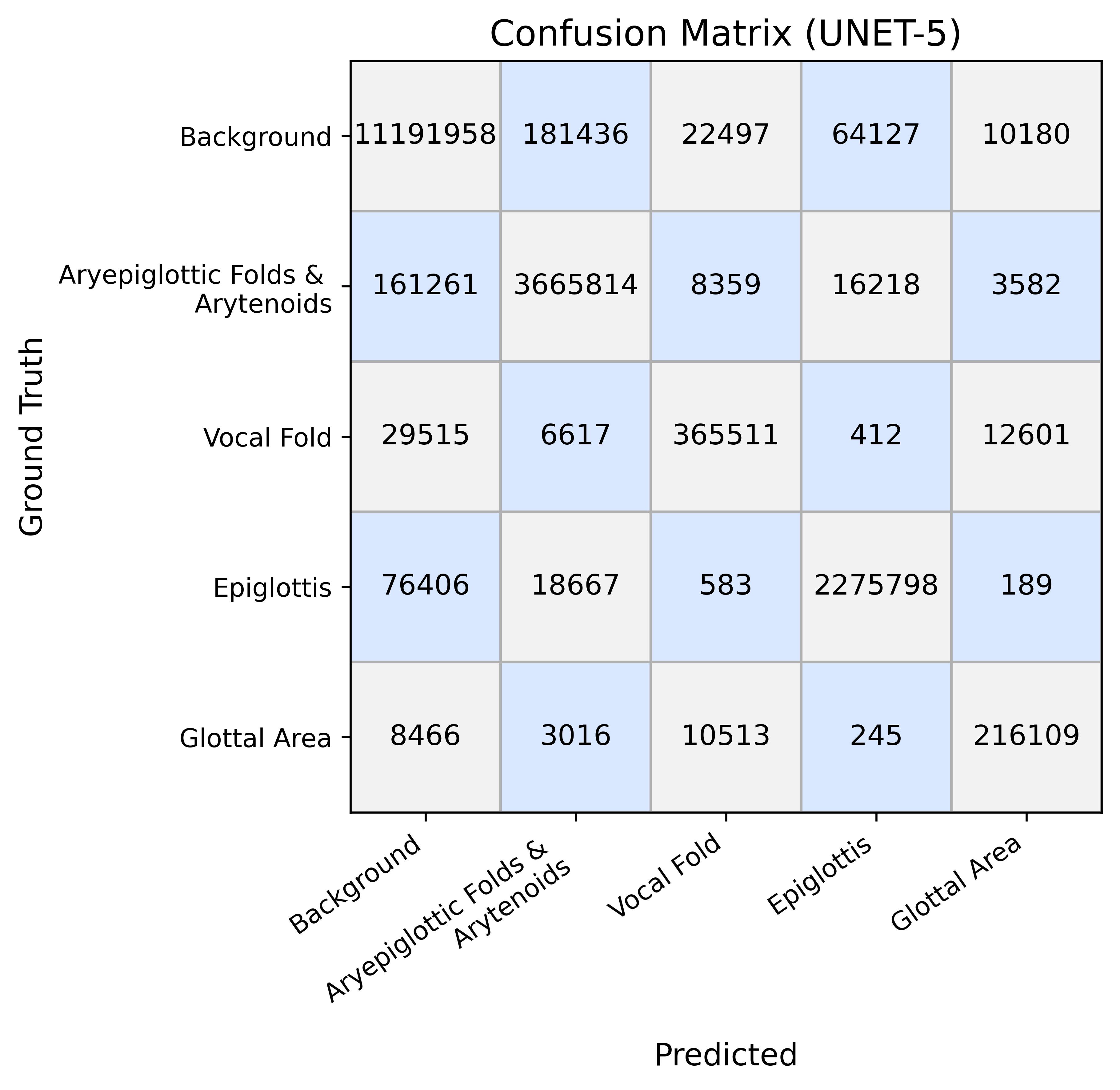}
    \label{fig:11b}
\end{subfigure}
\hfill
\begin{subfigure}[t]{0.32\columnwidth}
    \centering
    \caption{}
    \includegraphics[width=\columnwidth]{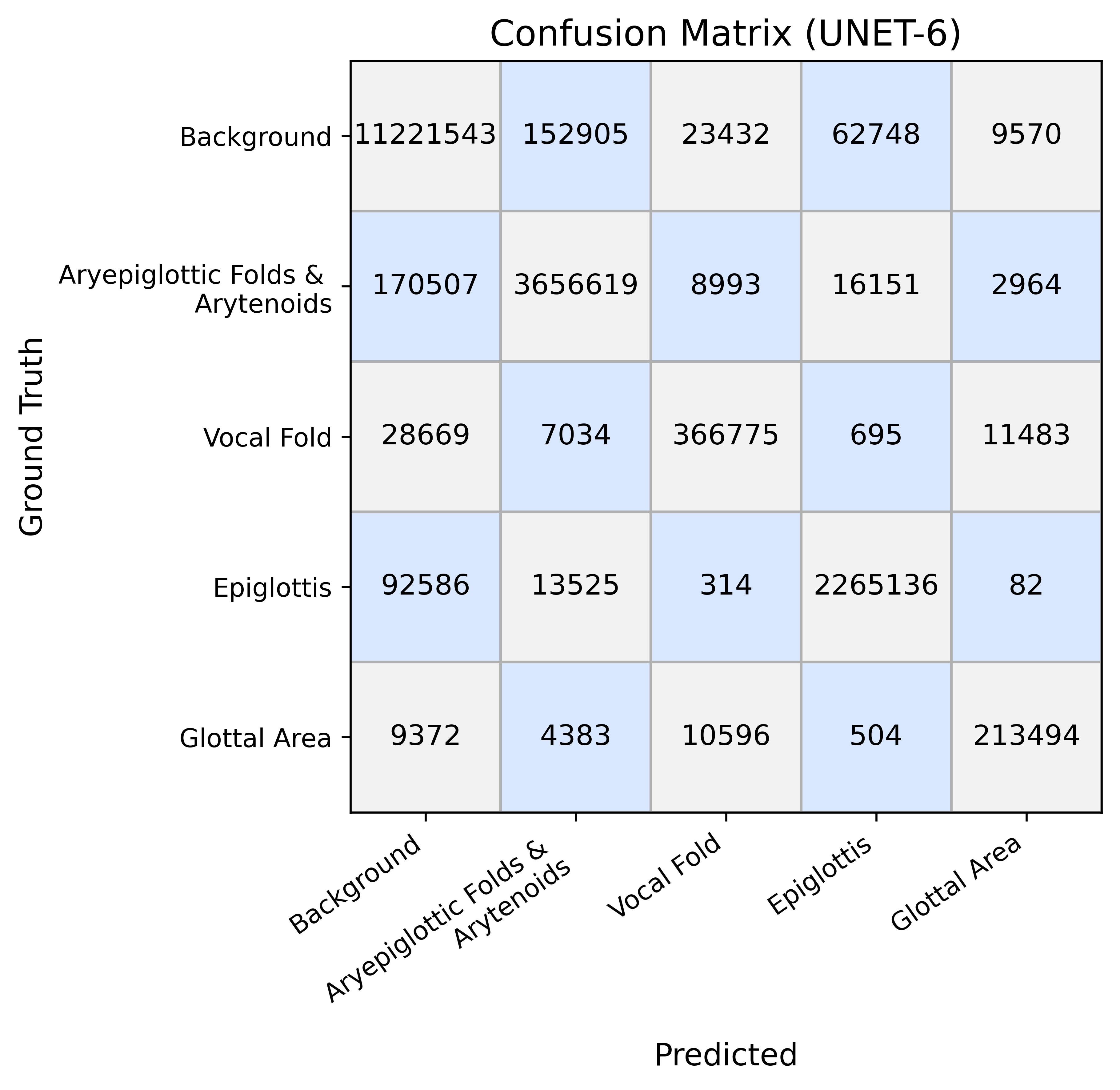}
    \label{fig:11c}
\end{subfigure}

\caption{Confusion matrices on the test set for (a) UNET-4, (b) UNET-5, and (c) UNET-6. Pixel counts along the rows correspond to the ground-truth labels for each class, while pixel counts along the columns correspond to the predicted labels.}
\label{f11}

\end{figure}

Figure~\ref{f11} presents the confusion matrices of the three U-Net models evaluated on the test set. The confusion matrices of all three U-Net models show strong diagonal elements, meaning that the number of true positives is much higher than the number of false positives and false negatives. This indicates strong segmentation performance across all five classes.
For both the predicted labels and the ground truth, the proportion of pixels from largest to smallest is: Background, Aryepiglottic Folds and Arytenoids, Epiglottis, Vocal Folds, and Glottal Area.
The training data also followed the same order. 
This imbalance in pixel distribution can introduce bias toward the more frequent classes. 
Overfitting further amplifies this effect, leading the model to assign more pixels to the common classes and fewer pixels to the less frequent classes.
The different performance metrics presented in the following sections can all be calculated from the confusion matrices, as described in Section~\ref{performance}.

For multiclass semantic segmentation, the IoU and Dice scores are presented in Table~\ref{T3} and Table~\ref{T4}, respectively.
For the Background class, all models achieve high Dice and IoU scores, whereas the scores for the Glottal Area and Vocal Folds classes are the lowest across all models.
This indicates that the Vocal Folds and Glottal Area are the most challenging regions to detect compared to the other anatomical zones. 
Although the performance of all models on these two classes is similar, UNET-5 identifies Glottal Area pixels slightly better, while it classifies Vocal Folds pixels slightly less accurately among the three models.
For the other anatomical zones, UNET-5 and UNET-6 perform almost similarly in terms of IoU and Dice scores, with UNET-4 showing slightly lower performance compared to the others.

\begin{table}[h]
\centering
\caption{IoU values for different U-Net models across segmentation classes.}
\label{T3}
\begin{tabular}{lccc}
\hline
\textbf{Class} & \textbf{UNET-4} & \textbf{UNET-5} & \textbf{UNET-6} \\ 
\hline
Background & 0.9463 & 0.9528 & 0.9533 \\
Aryepiglottic Folds and Arytenoids & 0.8866 & 0.9018 & 0.9067 \\
Vocal Folds & 0.8008 & 0.8005 & 0.8008 \\
Epiglottis & 0.9075 & 0.9279 & 0.9239 \\
Glottal Area & 0.8040 & 0.8158 & 0.8135 \\
\hline
\end{tabular}
\end{table}

\begin{table}[h]
\centering
\caption{Dice scores for different U-Net models across segmentation classes.}
\label{T4}
\begin{tabular}{lccc}
\hline
\textbf{Class} & \textbf{UNET-4} & \textbf{UNET-5} & \textbf{UNET-6} \\ 
\hline
Background & 0.9724 & 0.9759 & 0.9761 \\
Aryepiglottic Folds and Arytenoids & 0.9399 & 0.9484 & 0.9510 \\
Vocal Folds & 0.8894 & 0.8892 & 0.8894 \\
Epiglottis & 0.9515 & 0.9626 & 0.9604 \\
Glottal Area & 0.8914 & 0.8986 & 0.8971 \\
\hline
\end{tabular}
\end{table}

Table~\ref{T5} presents the class-wise accuracy (recall), while Table~\ref{T6} shows the overall accuracy. While recall measures how well the model avoids false negatives, precision measures how well the model avoids false positives. The precision values for the different models are presented in Table~\ref{T7}.
For the Background class, the false negative rate decreases with increasing network depth, while the false positive rate decreases from UNET-4 to UNET-5 and then increases slightly in UNET-6, although it remains lower than that of UNET-4.
For the Aryepiglottic Folds and Arytenoids, false negatives are lowest in UNET-5, whereas false positives continuously decrease from UNET-4 to UNET-6.
UNET-5 achieves the highest precision but the lowest recall for the Vocal Folds class, suggesting reduced false positives at the expense of increased false negatives. Consequently, this may result in under-segmentation of the vocal folds by UNET-5 compared to the other models.
In the Epiglottis class, recall increases in the order: UNET-4, UNET-6, and UNET-5. In contrast, precision almost increases as the network depth increases (with similar values in UNET-4 and UNET-5).
UNET-5 has the highest recall for the Glottal Area, with UNET-4 showing the second highest recall. However, the precision for this class increases with the number of network layers.
Consequently, when comparing the two models, UNET-5 and UNET-6, UNET-5 may slightly over-segment the Glottal Area, whereas UNET-6 may slightly under-segment it.

Referring to Table~\ref{T6}, the overall accuracy increases as the network depth increases. However, in multiclass segmentation, relying only on overall accuracy is not sufficient. Class-wise performance measures should be carefully considered when selecting a network, depending on which classes are more important for the application. In addition, the selection of a network should also consider training and inference speed, as well as the available computational resources. As network depth increases, training and prediction become slower and require more computational resources.

\begin{table}[h]
\centering
\caption{Recall values for different U-Net models across segmentation classes.}
\label{T5}
\begin{tabular}{lccc}
\hline
\textbf{Class} & \textbf{UNET-4} & \textbf{UNET-5} & \textbf{UNET-6} \\ 
\hline
Background & 0.9745 & 0.9757 & 0.9783 \\
Aryepiglottic Folds and Arytenoids & 0.9356 & 0.9509 & 0.9485 \\
Vocal Folds & 0.8922 & 0.8815 & 0.8845 \\
Epiglottis & 0.9474 & 0.9596 & 0.9551 \\
Glottal Area & 0.8981 & 0.9067 & 0.8957 \\
\hline
\end{tabular}
\end{table}

\begin{table}[h]
\centering
\caption{Overall test accuracy for different U-Net models.}
\label{T6}
\begin{tabular}{lc}
\hline
\textbf{Model} & \textbf{Accuracy} \\ 
\hline
UNET-4 & 0.9600 \\
UNET-5 & 0.9654 \\
UNET-6 & 0.9659 \\
\hline
\end{tabular}
\end{table}

\begin{table}[h]
\centering
\caption{Precision values for different U-Net models across segmentation classes.}
\label{T7}
\begin{tabular}{lccc}
\hline
\textbf{Class} & \textbf{UNET-4} & \textbf{UNET-5} & \textbf{UNET-6} \\ 
\hline
Background & 0.9704 & 0.9760 & 0.9739 \\
Aryepiglottic Folds and Aryetnoids & 0.9442 & 0.9459 & 0.9536 \\
Vocal Folds & 0.8866 & 0.8970 & 0.8943 \\
Epiglottis & 0.9556 & 0.9656 & 0.9658 \\
Glottal Area & 0.8848 & 0.8906 & 0.8986 \\
\hline
\end{tabular}
\end{table}

\section{Discussion} \label{discussion}
The current framework provides an automated segmentation approach for key laryngeal structures in HSV frames during production of connected speech, including the aryepiglottic folds and arytenoids, vocal folds, glottal area, and epiglottis.
While most existing studies focus primarily on detecting the glottal area (\citeauthor{i53}, \citeyear{i53}; \citeauthor{i54}, \citeyear{i54}; \citeauthor{i55}, \citeyear{i55}; \citeauthor{i56}, \citeyear{i56}; \citeauthor{i51}, \citeyear{i51}; \citeauthor{i52}, \citeyear{i52}), only a limited number have addressed multiclass segmentation involving the vocal folds in addition to the glottal area (\citeauthor{i57}, \citeyear{i57}; \citeauthor{res1}, \citeyear{res1}). 
\cite{res2} further included the epiglottis along with the vocal folds and glottal area in videolaryngoscopic recordings.
Studies that simultaneously detect the aryepiglottic folds and arytenoids along with the vocal folds and glottal area are currently lacking. Furthermore, the majority of existing studies do not include training datasets obtained from connected speech.
The inclusion of training labels from connected speech is essential, as it captures the complex and versatile movements of different laryngeal tissues, the dynamic behavior during transitional events, and disorder-related characteristics that are often revealed only in connected speech.
The present approach addressed these gaps by incorporating HSV training data from connected speech in addition to sustained vowel phonation and by detecting all major laryngeal structures visible in HSV frames.
Moreover, both normophonic and disordered subjects were included in the training dataset to enable the model to identify different tissues under varying dynamic conditions, orientations, and levels of supraglottal obstructions.

The training dataset used in this study was acquired from an HSV system equipped with a flexible nasolaryngoscope, which enabled the recording of connected speech.
However, HSV recordings acquired with a flexible nasolaryngoscope exhibit inferior image quality compared to those obtained with a rigid endoscope.
Different image processing steps, such as noise removal and histogram equalization, were applied to suppress noise and enhance the dynamic range of the images, allowing the deep learning model to more accurately separate multiple anatomical zones.
Despite implementing these processing steps, the image quality remained inferior to that of recordings obtained with rigid laryngoscopes, primarily mainly due to the limited dynamic range of the recording because of the low-light conditions.

We focused on developing a robust deep neural network architecture to accurately classify different anatomical zones in these images.
Convolutional neural networks that include dense layers may lose spatial relationships between pixels across different classes due to the flattening operation, which converts feature maps into one-dimensional vectors instead of preserving their two-dimensional structure.
To avoid this issue, we used a fully convolutional neural network, U-Net.
In U-Net, local (near-field) relationships are captured in the early layers where spatial resolution is high, while broader (far-field) contextual relationships are learned as the spatial resolution decreases in deeper layers.
Moreover, the skip connections between the encoder and decoder layers enable precise localization and preserve contextual information.
The model was trained using three different U-Net configurations with varying numbers of encoder and decoder layers to assess performance across different network depths.
Despite the use of lower-quality images, the model’s performance on the test set either exceeded that of existing models or achieved nearly comparable results.
In the studies of \cite{i57}, the Dice coefficients for the glottal area and vocal folds were reported as 0.85 and 0.91, respectively.
In our study, the corresponding highest Dice coefficients were 0.8986 for the glottal area using U-Net-5 and 0.8894 for the vocal folds using U-Net-4 and U-Net-6. 
\cite{res2} reported precision values of 0.962, 0.812, and 0.937 and recall values of 0.941, 0.685, and 0.642 for the epiglottis, glottal area, and vocal folds, respectively. In the current study, the highest precision values were 0.9658 (UNET-6), 0.8986 (UNET-6), and 0.8970 (UNET-5), while the highest recall values were 0.9596 (UNET-5), 0.9067 (UNET-5), and 0.8922 (UNET-4) for the same regions, respectively.

The test set included images from diverse normophonic and disordered subjects with various tissue orientations and configurations.
In addition to quantitative performance analysis, visual evaluation was performed on the test set to confirm that the results were both quantitatively reliable and qualitatively satisfactory.
The selection among the three U-Net networks implemented in this study depends on the specific segmentation priorities and the trade-off between performance and computational efficiency. U-Net-6 achieves the highest overall accuracy and generally strong class-wise performance, but requires greater computational resources and longer training and inference times. U-Net-5 provides comparable performance to U-Net-6 for several anatomical zones and may offer advantages for specific classes, while requiring less computation. U-Net-4 provides slightly lower segmentation performance overall but offers faster training and inference. Therefore, the choice among these networks should be guided by the anatomical structures of greatest interest, the desired segmentation performance, and the available computational resources.

The developed framework has strong potential for applications in voice disorder and voice dynamics research. Various quantitative measurements can be derived from the automatically detected zones to analyze the variability of dynamic behavior across different laryngeal tissues. These data can subsequently be used to quantify the extent of dynamic variation in different voice disorders relative to normal conditions.
Moreover, the model can be expanded to identify tissues responsible for supraglottal obstruction of the view of the vocal folds in laryngeal imaging.

While the network achieves accurate segmentation in most cases, incorrect and disconnected predicted regions are observed in some segmentations. These issues are planned to be addressed in future work by incorporating additional training images.
In addition, more advanced deep learning architectures may further improve the network’s performance.
Another issue with the network is its tendency to overpredict classes with larger proportions in the images and underpredict classes with smaller proportions. 
This issue may be addressed by employing strategies that better account for class imbalance and improve the representation of less prevalent classes during model training.

A major future implication of the network is the use of correctly predicted labels to expand the dataset for subsequent model development. This approach can serve as an automatic labeling tool, enabling the creation of larger annotated datasets for future studies.
In addition, this network can be readily applied to the detection of laryngeal structures in other types of laryngeal imaging beyond those used during training through transfer learning by freezing the encoder weights and updating only the decoder weights. This strategy would significantly reduce training time by decreasing the number of trainable parameters while maintaining high accuracy, as the network already captured essential features of laryngeal structures.

\section{Conclusions} \label{conclusions}
HSV enables the analysis of real intra-cycle variations in the motion of different laryngeal tissues.
Moreover, HSV recordings during running speech are particularly important for studying the non-stationary behavior of vocal fold vibrations and for characterizing disorder-specific motion patterns of multiple laryngeal tissues in disordered voices.
To address the challenges of manually analyzing the dynamic behavior of these tissues across thousands of HSV frames within short voiced segments, we developed an automated framework by implementing deep learning techniques.
Three U-Net–based networks with different layer depths were designed to detect and analyze important laryngeal landmarks, including the aryepiglottic folds and arytenoids, vocal folds, glottal area, and epiglottis.
The networks were trained on a dataset of labeled images from both disordered and normophonic subjects, including recordings of sustained vowel phonation and connected speech, to enable detection of these tissues under diverse dynamic conditions and orientations.
Several performance evaluation metrics were computed on unseen test images to assess the reliability and generalization of the trained models and to compare their performance.
This model is capable of automatically and reliably extracting many dynamic features of laryngeal landmarks, which are important for voice studies and clinical applications.

\section*{Acknowledgements}
The authors would like to acknowledge the support from the National Institutes of Health (NIH), National Institute on Deafness and Other Communication Disorders (NIDCD) under awards R21DC020003, K01DC017751, and R01DC019402, the U.S. Army Research Office (ARO) Young Investigator Program (YIP) under award W911NF-19-1-0444, the National Science Foundation (NSF) under award DMS-1923201, and Michigan State University Discretionary Funding Initiative. In addition, the authors would like to thank Dr. Stephanie R.C. Zacharias and Mayo Clinic for their help and support with the data collection, and the Institute for Cyber-Enabled Research (ICER) at Michigan State University for providing computational resources and facilities. Finally, the authors would like to acknowledge Alex Stewart and Bianca Imeraj for their help with the manual annotations.

\bibliography{citation}

\end{document}